\PassOptionsToPackage{dvipsnames}{xcolor}

\documentclass[runningheads]{llncs}
\usepackage{graphicx}

\usepackage{tikz}
\usepackage{comment}
\usepackage{amsmath,amssymb} % define this before the line numbering.
\usepackage{color}

\usepackage[accsupp]{axessibility}  % Improves PDF readability for those with disabilities.

\usepackage[capitalize]{cleveref}

\usepackage{xspace}
\newcommand{\LRenc}{LR-encoder\xspace}
\newcommand{\xb}{\mathbf{x}}
\newcommand{\yb}{\mathbf{y}}
\newcommand{\hb}{\mathbf{h}}
\newcommand{\lract}{\mathbf{g}}
\newcommand{\hract}{\hb}
\newcommand{\sract}{\Tilde{\hb}}
\newcommand{\zb}{\mathbf{z}}
\newcommand{\Ib}{\mathbf{I}}
\newcommand{\boldzero}{\mathbf{0}}
\newcommand{\Normal}{\mathcal{N}}

\usepackage[dvipsnames]{xcolor}

\usepackage[normalem]{ulem}
\useunder{\uline}{\ul}{}

\usepackage{booktabs}
\usepackage[normalem]{ulem}
\usepackage{multirow}

\begin{document}
% \linenumbers
\pagestyle{headings}
\mainmatter
\def\ECCVSubNumber{341}  % Insert your submission number here

\title{Image Super-Resolution With\\Deep Variational Autoencoders} % Replace with your title

%\begin{comment}
\titlerunning{Image Super-Resolution With Deep Variational Autoencoders}
\author{Darius Chira\inst{1}\thanks{Equal contribution, alphabetical order. 
Contact: \textless darius.chira14@gmail.com\textgreater\ and \textless ilianbg9@gmail.com\textgreater.} 
\and
Ilian Haralampiev\inst{1}\textsuperscript{$\star$} \and
Ole Winther\inst{1,2,3}\and \\
Andrea Dittadi\inst{1,4}\thanks{Equal advising, alphabetical order.
\\[5pt]\hspace*{-10pt}ECCV 2022 Workshop on Advances in Image Manipulation%
} \and
Valentin Liévin\inst{1}\textsuperscript{$\star\star$}}
\authorrunning{D. Chira, I. Haralampiev, et al.}
\institute{%
Technical University of Denmark\\
\and
University of Copenhagen\\ 
\and
Rigshospitalet, Copenhagen University Hospital\\
\and
Max Planck Institute for Intelligent Systems, Tübingen, Germany
}
%\end{comment}
%******************
\maketitle
\begin{abstract}
Image super-resolution (SR) techniques are used to generate a high-resolution image from a low-resolution image. Until now, deep generative models such as autoregressive models and Generative Adversarial Networks (GANs) have proven to be effective at modelling high-resolution images. VAE-based models have often been criticised for their feeble generative performance, but with new advancements such as VDVAE, there is now strong evidence that deep VAEs have the potential to outperform current state-of-the-art models for high-resolution image generation. In this paper, we introduce VDVAE-SR, a new model that aims to exploit the most recent deep VAE methodologies to improve upon the results of similar models. VDVAE-SR tackles image super-resolution using transfer learning on pretrained VDVAEs. The presented model is competitive with other state-of-the-art models, having comparable results on image quality metrics.
\end{abstract}

\section{Introduction}

\label{sec:intro}

Single Image Super-Resolution (SISR) consists in producing a high-resolution image from its low-resolution counterpart. Image super-resolution has long been considered one of the most arduous challenges in image processing. This is yet another computer vision task that was transformed by the deep learning revolution and has potential applications including but not limited to medical imaging, security, computer graphics, and surveillance.

Deep generative models have been shown to excel at image generation. This is particularly true for autoregressive models \cite{nade,pixel_rnn,pixel_cnn,pixel_snail,image-transformer} and Generative Adversarial Networks (GAN) \cite{goodfellow2014generative,zhu2017unpaired,karras2019style,brock2018large}, whereas Variational Autoencoders (VAE) \cite{kingma2013auto,rezende2014stochastic} have long been thought to be unable to produce high-quality samples. However, recent improvements in VAE design, such as using a hierarchy of latent variables and increasing depth \cite{kingma2016improved,maaloe2019biva,vahdat2020nvae,child2020very} have demonstrated that deep VAEs can compete with both GANs and autoregressive models for high-resolution image generation.
The current state-of-the-art VAE is the Very Deep Variational Autoencoder (VDVAE) \cite{child2020very} which successfully scales to 78 stochastic layers, whereas previous work only experimented with up to 40 layers \cite{vahdat2020nvae}. 

Since the VAE is an unconditional generative model, in order to perform image super-resolution it has to be turned into a \emph{conditional} generative model which generates data depending on additional conditioning data. This can be achieved by using the framework of Conditional Variational Autoencoders (CVAE) \cite{NIPS2015_8d55a249}, where the prior is conditioned on an additional random variable and parameterized by a neural network.
In this work, we introduce \textbf{VDVAE-SR}, a VDVAE conditioned on low-resolution images by adding a new component that we call \LRenc as it resembles the encoder of the original VDVAE. This component is connected to the decoder, passing information on each layer in the top-down path both to the prior and the approximate posterior. During training, the latent distributions of the low- and high-resolution images are matched using the KL divergence term in the evidence lower bound (ELBO). The learned information is used in generative mode, where only the low-resolution image is included in the model.

A drawback of deep models such as the VDVAE is that they require a large amount of computing and training time. One way to compensate in that regard is to apply transfer learning and utilize a pre-trained model in order to speed up the process. However, this is not always straightforward in practice as presenting a pre-trained model with new data could lead to exploding gradients. This is particularly relevant for deep variational autoencoders as they are prone to unstable training and can be sensitive to hyperparameters changes. We show that using transfer learning for such a model is possible, and we describe the methods to do so, by making only certain parts of the network trainable and using gate parameters to stabilise the process.

We fine-tune a VDVAE model pretrained on FFHQ 256x256 \cite{karras2019style} using DIV2K \cite{Agustsson_2017_CVPR_Workshops}, a common dataset in the image super-resolution literature \cite{lim2017enhanced,wang2018esrgan,niu2020single,Dai_2019_CVPR}.
We evaluate the fine-tuned model on a number of common datasets in the literature of single image super-resolution: Set5 \cite{BMVC.26.135}, Set14 \cite{zeyde2010single}, Urban100 \cite{Huang_2015_CVPR}, BSD100 \cite{937655}, and Manga109 \cite{Matsui_2016}. Following previous work \cite{ledig2017photo,wang2018esrgan,niu2020single}, we test our approach both quantitatively, in terms of PSNR and SSIM metrics, and qualitatively, by visually inspecting the generated images, and compare our results against three state-of-the-art super-resolution methods: EDSR \cite{lim2017enhanced}, ESRGAN \cite{wang2018esrgan} and RFANet \cite{9156371}.
We investigate the role of the sampling temperature, which controls the variance of samples at each stochastic layer in VDVAEs, and show results generated with low and high temperatures. By sampling with a lower temperature, the model achieves quantitative scores better than ESRGAN, but slightly lower than EDSR. At the same time, qualitatively, when sampling with a higher temperature, the images look sharper and less blurred than those generated by the EDSR model. Even though, in general, ESRGAN generates sharper images, it is prone to produce more artifacts as well. We believe that our proposed method shows a good compromise between visual artifacts and image sharpness.

We summarize our contributions as follows:
\begin{enumerate}
\item We propose VDVAE-SR, an adaptation of very deep VAEs (VDVAEs) for the task of single image super-resolution. VDVAE-SR introduces an additional component that we call \LRenc, which takes the low-resolution image as input, while its output is used to condition the prior.

\item  We show how to utilize transfer learning and achieve stable training in order to take advantage of a VDVAE model already pre-trained on 32 V100 GPUs for 2.5 weeks.

\item  We present competitive qualitative and quantitative results compared to state-of-the-art methods on popular test datasets for 4x upscaling.

\end{enumerate}

\section{Related Work}
One of the first successes in image super-resolution is the SR-CNN \cite{dong2015image}, which is based on a three-layer CNN structure and uses a bicubic interpolated low-resolution image as input to the network. Later, with the proposal of residual neural networks (ResNets) \cite{he2016deep}, which provide fast training and better performance for deep architectures, numerous works have adapted ResNets-based models to the task of super-resolution, such as SR-ResNet \cite{ledig2017photo} and SR-DenseNet \cite{tong2017image}. One of the frequently used CNN-based super-resolution models in comparative studies is EDSR \cite{lim2017enhanced}, where the authors use ResNets without batch normalization in the residual block, achieving impressive results and getting first place on the NTIRE2017 Super-Resolution Challenge.
 
In terms of GAN-based image super-resolution models, several methods have gained a lot of popularity starting with SRGAN \cite{ledig2017photo} where the authors argue that most popular metrics (PSNR, SSIM) do not necessarily reflect perceptually better SR results and that is why they use an extensive mean opinion score (MOS) for evaluating perceptual quality. With that in mind, SRGAN introduces a perceptual loss different from previous work, based on adversarial as well as content loss. Another method, ESRGAN \cite{wang2018esrgan}, builds upon SRGAN by improving the network architecture removing all batch normalization layers and introducing a new Residual in Residual Dense Block (RRDB). In addition, an enhanced discriminator is used based on Relativistic GAN \cite{jolicoeur2018relativistic} and the features before the activation loss are used to improve perceptual loss. 

A recent work that uses VAEs for image super-resolution is the srVAE \cite{gatopoulos2020super}, which consists of a VAE with three latent variables, one of them being a downscaled version of the original image. This work shows impressive generative performance in terms of FID score when tested on ImageNet-32 and CIFAR-10, but no quantitative results of their super-resolution model are reported. Another recent work that uses a VAE-based model for image super-resolution is VarSR \cite{hyun2020varsr}. This work focuses on very low-resolution images (8x8) and shows better results compared to some popular super-resolution methods.

Deep VAEs such as \cite{kingma2016improved,maaloe2019biva,vahdat2020nvae,child2020very} adapt their architecture from Ladder VAEs (LVAE) \cite{sonderby2016ladder}, which introduce a novel top-down inference model and achieve stable training with multiple stochastic layers. A method that improved upon the LVAE is the Bidirectional-Inference VAE (BIVA) \cite{maaloe2019biva} adding a deterministic top-down path in the generative model and applying a bidirectional inference network. These modifications solved the variable collapse issue of the LVAE which may occur when the architecture consists of a very deep hierarchy of stochastic latent variables. Recently, NVAE \cite{vahdat2020nvae} reported further improvements by using normalizing flows in order to allow for more expressive distributions and thus outperform the state-of-the-art among non-autoregressive and VAE models. Finally, the VDVAE model \cite{child2020very} demonstrated that the number of stochastic layers matters greatly for performance, achieving better results than previous VAE-based models and some autoregressive ones, having the potential to outperform those as well.

Denoising Diffusion Probabilistic Models (DDPM) \cite{ddpm} are the latest addition to the family of probabilistic generative models. DDPMs define a diffusion process that progressively turns the input image into noise, and learn to synthesize images by inverting that process. DDPMs and variations thereof excel at high-resolution image generation \cite{nichol2021improved,dhariwal2021diffusion} and have been successfully applied to the task of single image super-resolution \cite{cascaded_ddpm,sr_diff}.

\section{Preliminaries}

In this section, we define variational autoencoders (VAEs) and conditional VAEs (CVAEs), for which we derive the evidence lower bound. We then introduce the VDVAE using the VAE framework.

\subsection{Variational Autoencoders}

The Variational Autoencoder \cite{kingma2013auto} is a generative model built on probabilistic principles. It consists of a joint model $p_\theta(\xb, \zb) = p_\theta(\xb|\zb) p_\theta(\zb)$ parameterized by $\theta$ and an approximate posterior $q_\phi(\zb|\xb)$ parameterized by $\phi$. All models are implemented using neural networks. During generation, the latent variable $\mathbf{z}$ is sampled from the prior and the observation variable $\xb$ is sampled from the observation model following $ \zb  \sim p_{\theta} (\zb), \xb \sim p_{\theta} (\xb|\zb) $.

VAE models are optimized with stochastic gradient ascent to maximize the marginal likelihood:
\begin{equation}
p_\theta(\xb) = \int_{\zb} p_\theta(\xb|\zb) p_\theta(\zb) d \zb
\end{equation}
In practice, $p_\theta(\xb)$ is intractable of the integration over $\zb$, which makes the posterior $p_\theta(\zb|\xb)$ also intractable. Variational Inference (VI) solves the intractability of $p_\theta(\zb|\xb)$ using an approximate posterior $q_\phi(\zb|\xb)$.
The resulting objective function, the evidence lower bound (ELBO), is further derived using Jensen's inequality and expressed as:
\begin{equation}
\mathcal{L}(\xb ; \theta, \phi) = E_{q_{\phi}(\zb|\xb)}  \left[log \frac{p_{\theta}(\xb, \zb)}{q_\phi(\zb | \xb)} \right] \leq \log p_\theta(\xb) \ .
\label{eq:vae_elbo}
\end{equation}

\subsection{Conditional Variational Autoencoders}

In order to generate specific data as in the case of image super-resolution, where we need to generate a high-resolution image from its low-resolution counterpart, the Conditional Variational Autoencoders (CVAE) can be used.

Similar to the VAE, the CVAE is also built on probabilistic principles. CVAE is optimized to maximize the marginal probability similar to \cref{eq:vae_elbo} but this time conditioned on a random variable which could be for example a low-resolution image $\yb$:
\begin{equation} \label{eq:CVAE_gen}
 p_\theta (\xb|\yb) = \int_\zb p_\theta(\xb | \yb,\zb) p(\zb|\yb) d \zb \
\end{equation}
The posterior of the latent variables is: 
\begin{equation} \label{eq:CVAE_bayes}
p_{\theta}(\zb|\xb,\yb) = \frac{p_{\theta}(\xb|\zb,\yb)p_{\theta}(\zb|\yb)}{p_{\theta}(\xb|\yb)}
\end{equation}
where again $p_{\theta}(\xb|\yb)$ is intractable and needs to be approximated using a variational distribution $q_{\phi}(\zb|\xb,\yb) \approx p_{\theta}(\zb|\xb,\yb)$.
The conditional ELBO for the CVAE can be derived again using Jensen's inequality, resulting in:

\begin{equation}
\mathcal{L}(\xb , \yb; \theta, \phi) = E_{q_{\phi}(\zb|\xb, \yb)}  \left[\log \frac{p_{\theta}(\xb, \zb, \yb)}{q_\phi(\zb | \xb, \yb)} \right] \leq \log p_\theta(\xb | \yb) \ .
\label{eq:ELBO_CVAE}
\end{equation}

\subsection{Very Deep Variational Autoencoder (VDVAE)}

The VDVAE \cite{child2020very} 
consists of a hierarchy of layers of latent variables conditionally dependent on each other. This results in a more flexible prior and posterior compared to a simple diagonal Gaussian prior which could be too limiting. An iterative interaction between ``bottom-up'' and ``top-down'' layers is achieved through parameter sharing between the inference and generative models in each layer. The prior and the approximate posterior for a model with $K$ stochastic layers factorize as:
\begin{align}
p_{\theta}(\zb) &= p_{\theta}(\zb_0)p_{\theta}(\zb_1|\zb_0)...p_{\theta}(\zb_K|\zb_{<K})\\
q_{\phi}(\zb|\xb) &= q_{\phi}(\zb_0|\xb)q_{\phi}(\zb_1|\zb_0,\xb)...q_{\phi}(\zb_K|\zb_{<K},\xb)
\end{align}
where $p_{\theta}(\zb_0)$ is a diagonal Gaussian distribution $\Normal(\zb_0\,|\,\boldzero,\Ib)$ and the latent variable group $\zb_0$ is at the top layer that corresponds to small number of latent variables at low resolution. Intuitively, $\zb_K$ is at the bottom of the network having a larger number of latent variables at high resolution. 

\begin{figure}
\centering
\includegraphics[height=6cm]{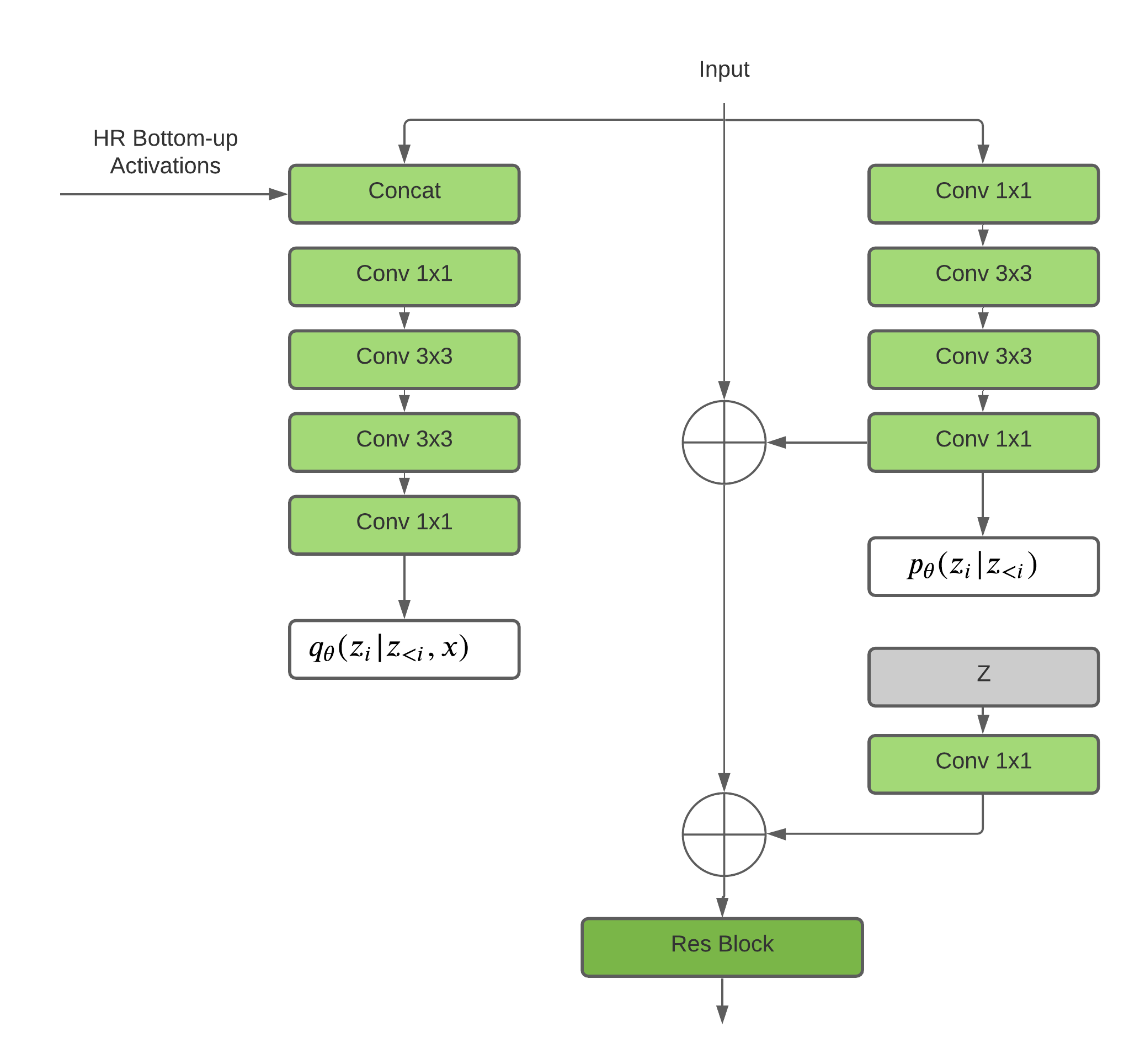}
\caption{Top-down block of the VDVAE \cite{child2020very}.}
\label{fig:VDVAE_top_down_block}
\end{figure}

The VDVAE architecture is composed of blocks of two types: the residual blocks (bottom-up path) and the top-down blocks (\cref{fig:VDVAE_top_down_block}). The top-down blocks are also residual and handle two tasks: processing the information flowing through the decoder and handling the stochasticity. Each top-down block of index $j > 0$ handles the distributions $q_\phi(\zb_{j}|\zb_{j-1}, \xb)$ and $p_\theta(\zb_{j} | \zb_{j-1})$. Top-down blocks are composed sequentially. Therefore, we can define $\hract_j$ as the input to the top-down block of index $j$, where $\hb_j$ is a function of the samples $\zb_{<j}$. This allows us to express the VDVAE model as:
\begin{equation}\label{eq:vdvae}
     p_{\theta}(\zb) = p_{\theta}(\zb_0) \prod_{j = 1}^{K} p_{\theta}(\zb_j| \hract_j), 
     \qquad q_{\phi}(\zb|\xb) = q_{\phi}(\zb_0|\xb) \prod_{j = 1}^{K} q_{\phi}(\zb_j| \hract_j, \xb) \ . \\ 
\end{equation}

\section{VDVAE-SR}\label{sec:vdvae-sr}
In this section, we introduce the proposed VDVAE-SR model. We provide an overview of the model architecture, after which we detail the conditional prior network and its integration with the VDVAE model.
\begin{figure}
\centering
\includegraphics[height=11cm]{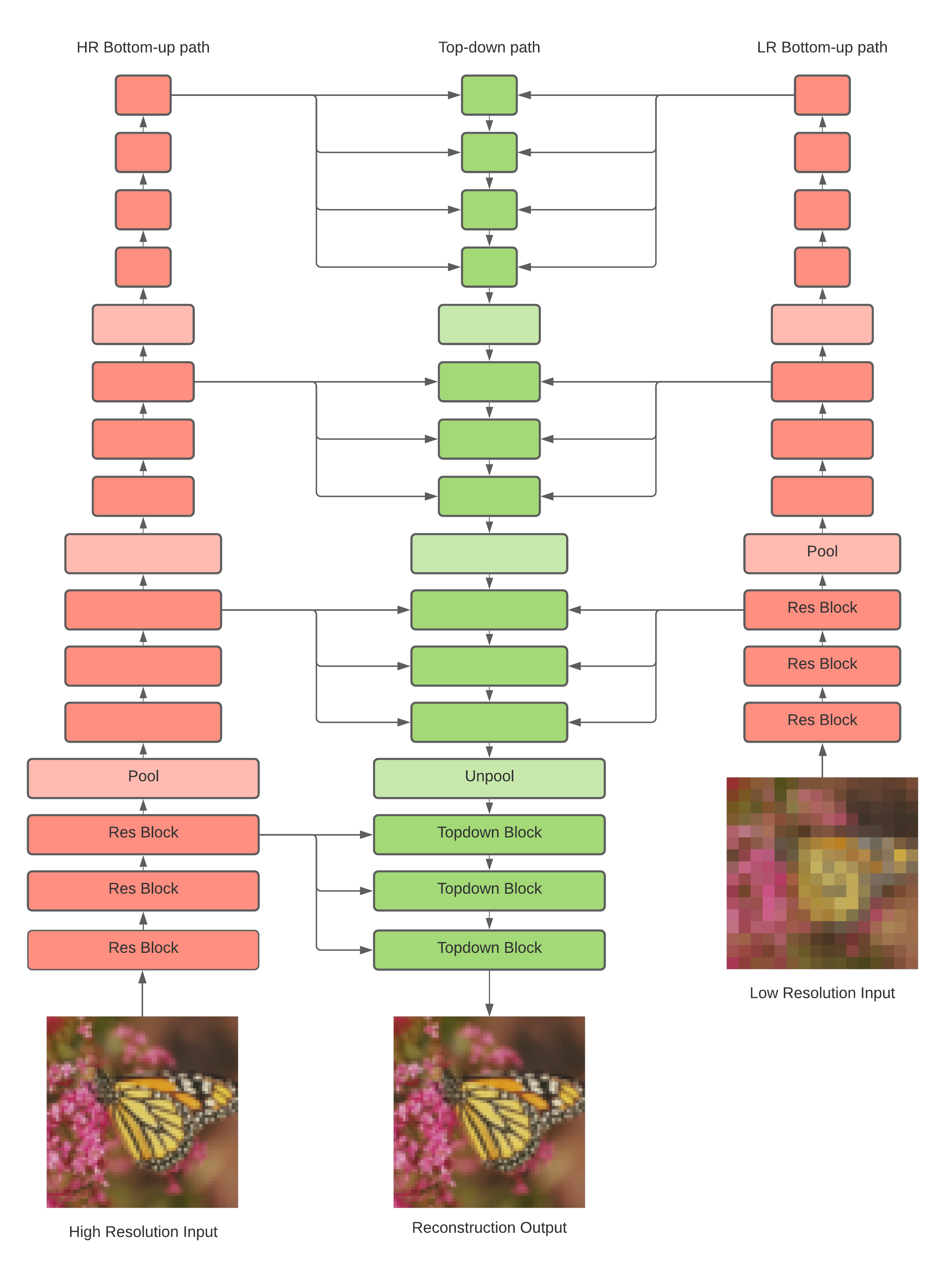}
\caption{Network Architecture of the proposed VDVAE-SR Model.}
\label{fig:VDVAE-SR_architecture}
\end{figure}

\subsection{\LRenc}
The dependency on the lower-resolution image $\yb$ is implemented using the encoder of a lower-resolution VDVAE of depth $K' < K$, which we call \LRenc. The \LRenc maps the lower-resolution image to latent space, providing one activation $\lract_j$ for each layer $j \in [0, K']$. Each activation $\lract_j$ is defined as the output of the bottom-up residual block of index $j$.

\subsection{Conditional Prior}
\begin{figure}
\centering
\includegraphics[height=8cm]{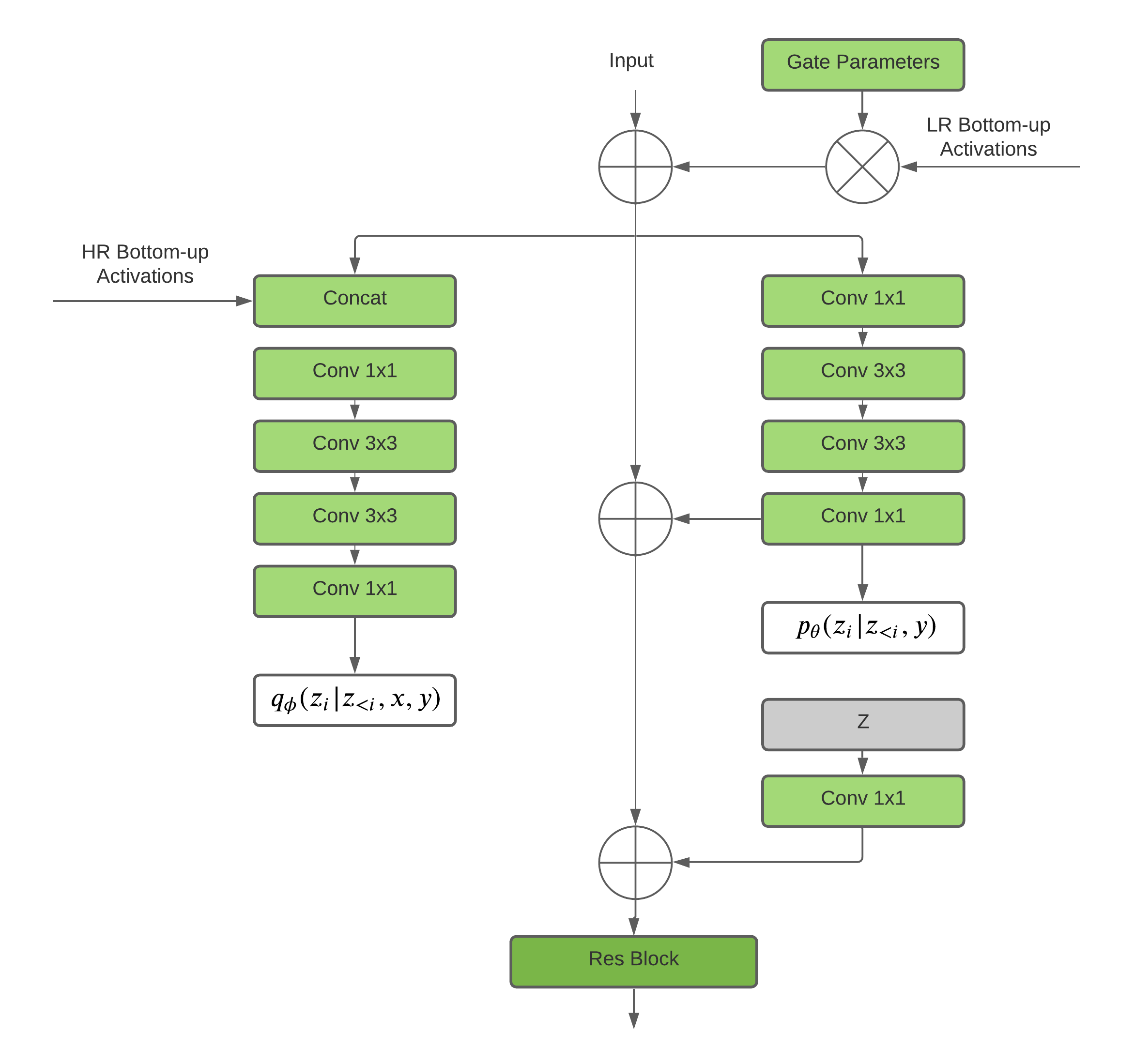}
\caption{Top-down block of the VDVAE-SR.}
\label{fig:VDVAE-SR_top_down_block}
\end{figure}
The top-down path, or decoder, of the VDVAE is modified to depend on $\yb$ using the \LRenc activations $\lract_0, \ldots, \lract_{K'}$. This results in a conditional prior $p{_\theta}(\zb|\yb)$ that maps the low-resolution image $\yb$ to a distribution over the latent variables $\zb$ (\cref{fig:VDVAE-SR_architecture}).

The architecture of the VDVAE-SR is identical to the one of the VDVAE, except for two alterations:
\begin{enumerate}

    \item The input to each top-down block (see \cref{fig:VDVAE-SR_top_down_block}) is defined as:
    \begin{equation}
      \sract_j =
      \begin{cases}
        \lract_j & \text{if $j = 0 $} \\
        \hract_j + \alpha_j \lract_j & \text{if $j \in [1, K']$} \\
        \hract_j & \text{otherwise}
      \end{cases}
    \end{equation}
    where $\alpha_1, \ldots, \alpha_{K'}$ are scalar gate parameters initialized to zero \cite{bachlechner2020rezero}.

    \item The top layer is conditioned on $\yb$ such that 
    \begin{equation}
        p_{\theta}(\zb_0|\sract) = \Normal\left(\zb_0 \, | \, \mu_\theta(\sract_0),\, \sigma_\theta(\sract_0)\right) \ ,
    \end{equation}
    where $\mu_\theta$ and $\sigma_\theta$ are linear layers mapping the output of the top-most \LRenc layer to the parameter-space of $p_\theta(\zb_0 | \yb)$. 
    
\end{enumerate}

\subsection{Generative model and inference network}
Because of the sharing of the top-down model between the generative model and the inference network \cite{sonderby2016ladder}, the conditional inference network naturally arises from the alteration of the prior, without further modification. Using the activations $\sract_0, \ldots, \sract_{K}$ and the definition of the VDVAE given in \cref{eq:vdvae}, we define the VDVAE-SR as:

\begin{equation}\label{eq:vdvae-sr}
     p_{\theta}(\zb|\yb) = \prod_{j = 0}^{K} p_{\theta}(\zb_j| \sract_j), 
     \qquad q_{\phi}(\zb|\yb,\xb) = q_{\phi}(\zb_0|\xb) \prod_{j = 1}^{K} q_{\phi}(\zb_j| \sract_j, \xb) \ . \\ 
\end{equation}

\section{Experiments}

\subsection{Datasets}
\subsubsection{Training Dataset.}
We train our models on the DIV2K dataset, introduced by \cite{Agustsson_2017_CVPR_Workshops}. The DIV2K dataset consists of 800 RGB high-definition high-resolution images for training, 100 images for validation, and 100 for testing. The dataset contains a variety of diverse pictures, including different types of shot such as portrait, scenery, and object shots.

\subsubsection{Test Datasets.}
We test our method on popular benchmarking datasets commonly used in single-image super resolution: Set5 \cite{BMVC.26.135}, Set14 \cite{zeyde2010single}, Urban100 \cite{Huang_2015_CVPR}, BSD100 \cite{937655}, and Manga109 \cite{Matsui_2016}. Having multiple test datasets gives a better understanding of the strengths and shortcomings of our model, since these datasets contain different types of pictures: BSD100, Set5, and Set14 mostly consist of natural images with a broad range of styles, while the focus on Urban100 is mainly on buildings and urban scenes, and Manga109 consists of drawings of Japanese manga.

\subsection{Implementation Details}

Since it takes about 2.5 weeks to train a VDVAE model on FFHQ 256x256 on 32 NVIDIA V100 GPUs, we choose to rely on pretrained VDVAEs and adapt them to the super-resolution task. We use a pretrained VDVAE with a stochastic depth of 62 layers. Our method, VDVAE-SR, includes the original VDVAE encoder and decoder, which we initialize with the weights from the pretrained model. We then freeze the encoder, allow fine-tuning of the decoder, and train the \LRenc from scratch. We optimize the model end-to-end for 100,000 steps using the Adam optimizer \cite{kingma2014adam} with a learning rate of $5\cdot10^{-4}$ and batch size of 1 on one NVIDIA V100 GPU.

When using transfer learning, it was observed that the model suffered from exploding gradients if the new information from the \LRenc was introduced in an uncontrolled manner. Introducing gate parameters similar to the approach in \cite{bachlechner2020rezero} significantly improved training stability.

\subsection{Evaluation}

In terms of evaluation metrics, we use the traditional PSNR and SSIM quality measures, both widely used as metrics for image restoration tasks. While PSNR (Peak Signal to Noise Ratio) is calculated based on the mean squared error of the pixel-to-pixel difference, the SSIM (Structural Similarity Method) is considered to have a closer correlation with human perception by calculating distortion levels based on comparisons of structure, luminance, and contrast. Additional to the traditional PSNR and SSIM metrics, we evaluate the produced images using the DISTS \cite{ding2020iqa} score, which has showed evidence that the metric matches closer to human perception.
We quantitatively evaluate different super-resolution methods by applying them to low-resolution images and computing the PSNR, SSIM and DISTS metrics using the super-resolution output and the reference high-resolution image.
For the PSNR and SSIM, all pictures are converted from RGB to YCbCr and the metrics are computed on the Y channel (luma component) of the pictures. The reason for this is that it has been observed (e.g., in \cite{Pisharoty}) that the results of evaluating on the luminosity channel in the YCbCr color space, rather than on the usual RGB representation, are closer to the actual perceived structural noise of the image. We thus adopt the same approach, following prior work. Finally, note that the YCbCr space is used during the testing phase exclusively, while the training and validation are still performed in the RGB color space.\looseness=-1

\subsection{Results}

\subsubsection{Quantitative Results.}

We compare our method to three other super-resolution methods, namely EDSR, ESRGAN and RFANet, based on their official implementation. The quantitative results on PSNR and SSIM are shown in \cref{tab:eval_metrics}, where EDSR performs best on both metrics, with our method ($t = 0.1$) closely following on second place.

\begin{table}
\centering
\caption{Evaluation metrics using PSNR and SSIM on the Y channel and DISTS. The number next to VDVAE-SR (our method) denotes the temperature used for sampling. The best scores are represented in \textbf{bold}, while the second best results are {\ul underlined}.}
\resizebox{\columnwidth}{!}{%
\begin{tabular}{cccccccccccclccc}
\toprule
\multicolumn{1}{l}{\multirow{2}{*}{\textbf{Dataset}}} & \multicolumn{3}{c}{\textbf{EDSR}}                           & \multicolumn{3}{c}{\textbf{ESRGAN}}       & \multicolumn{3}{c}{\textbf{RFANet}}                         & \multicolumn{3}{c}{\textbf{VDVAE-SR 0.1}} & \multicolumn{3}{c}{\textbf{VDVAE-SR 0.8}} \\ \cmidrule(lr){2-4} \cmidrule(lr){5-7} \cmidrule(lr){8-10} \cmidrule(lr){11-13} \cmidrule(lr){14-16}
\multicolumn{1}{l}{}                                  & PSNR           & SSIM           & \multicolumn{1}{l}{DISTS} & PSNR  & SSIM  & \multicolumn{1}{l}{DISTS} & PSNR           & SSIM           & \multicolumn{1}{l}{DISTS} & PSNR           & SSIM           & DISTS   & PSNR  & SSIM  & \multicolumn{1}{l}{DISTS} \\ \midrule
Set5                                                  & {\ul 31.97}    & {\ul 0.902}    & 0.121                     & 30.39 & 0.864 & \textbf{0.078}            & \textbf{32.53} & \textbf{0.908} & 0.119                     & 31.48          & 0.886          & 0.123   & 30.51 & 0.869 & {\ul 0.108}               \\
Set14                                                 & \textbf{28.33} & \textbf{0.800} & 0.097                     & 26.20 & 0.720 & \textbf{0.064}            & 27.33          & 0.774          & {\ul 0.092}               & {\ul 27.99}    & {\ul 0.776}    & 0.105   & 27.62 & 0.761 & 0.097                     \\
BSDS100                                               & \textbf{28.46} & \textbf{0.781} & 0.158                     & 25.87 & 0.690 & \textbf{0.094}            & 27.04          & {\ul 0.758}    & 0.154                     & {\ul 28.05}    & 0.752          & 0.169   & 27.69 & 0.738 & {\ul 0.152}               \\
Manga109                                              & \textbf{30.85} & \textbf{0.918} & \textbf{0.009}            & 28.77 & 0.870 & {\ul 0.010}               & 21.09          & 0.739          & 0.015                     & {\ul 29.92}    & {\ul 0.904}    & 0.013   & 29.55 & 0.896 & 0.011                     \\
Urban100                                              & {\ul 26.02}    & {\ul 0.798}    & 0.029                     & 24.36 & 0.748 & {\ul 0.024}              & \textbf{26.89} & \textbf{0.823} & \textbf{0.023}            & 25.36          & 0.759          & 0.037   & 25.15 & 0.750 & 0.034                     \\ \bottomrule
\end{tabular}%
}
\label{tab:eval_metrics}
\end{table}

As first discussed in \cite{ledig2017photo}, the PSNR and SSIM scores tend to favor smoother images, this being attributed to the nature of how these metrics are calculated, which is in contrast to human visual perception. This is confirmed by the obtained scores of our method using different temperatures as decreasing the variance produces more averaged-out images and thus higher scores. Based on the DISTS metric, ESRGAN performs best on three datasets. Our method with higher temperature follows on second place on the Set5 and BSDS100 datasets.

\subsubsection{Qualitative Results.}

\begin{figure}
\centering
\includegraphics[height=4.2cm]{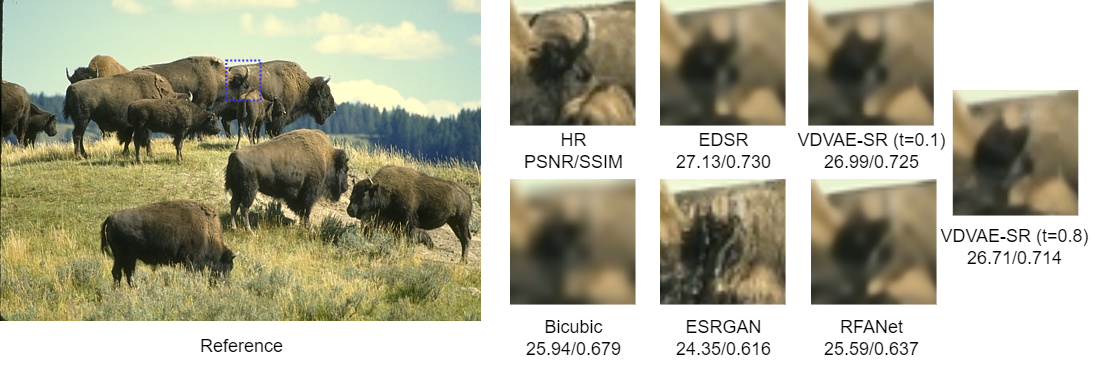}
\caption{SR output comparison between multiple models for a picture (image 376043) of the BSD100 dataset.}
\label{fig:bulls_comp}
\end{figure}

\begin{figure}
\centering
\includegraphics[height=4.5cm]{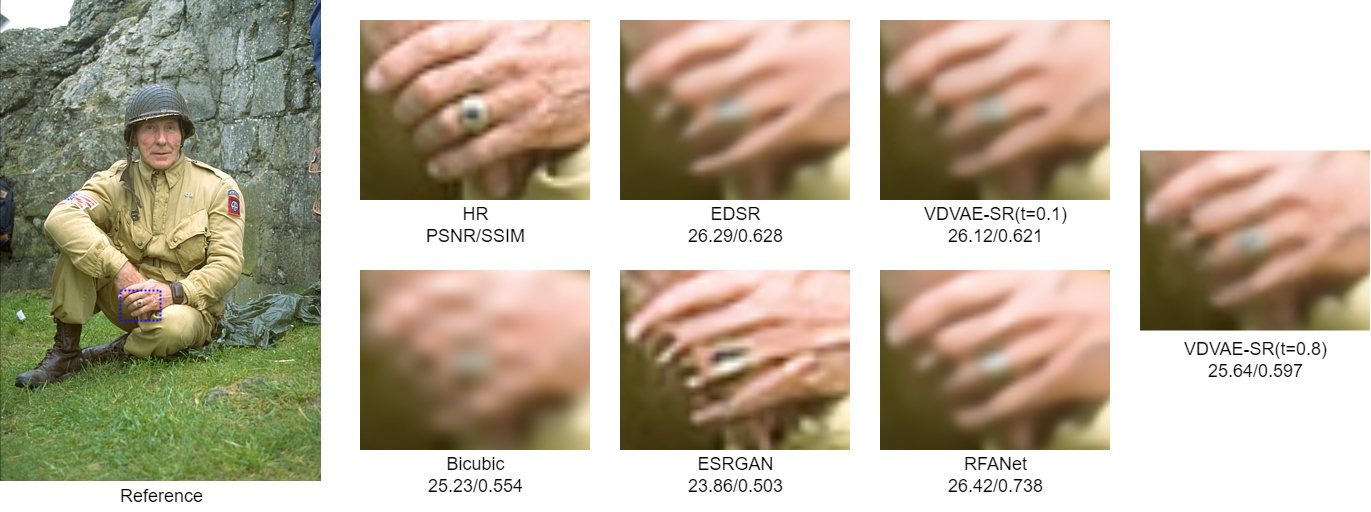}
\caption{SR output comparison between multiple models for a picture (image 38092) of the BSD100 dataset.}
\label{fig:soldier_comp}
\end{figure}

\begin{figure}
\centering
\includegraphics[height=5.5cm]{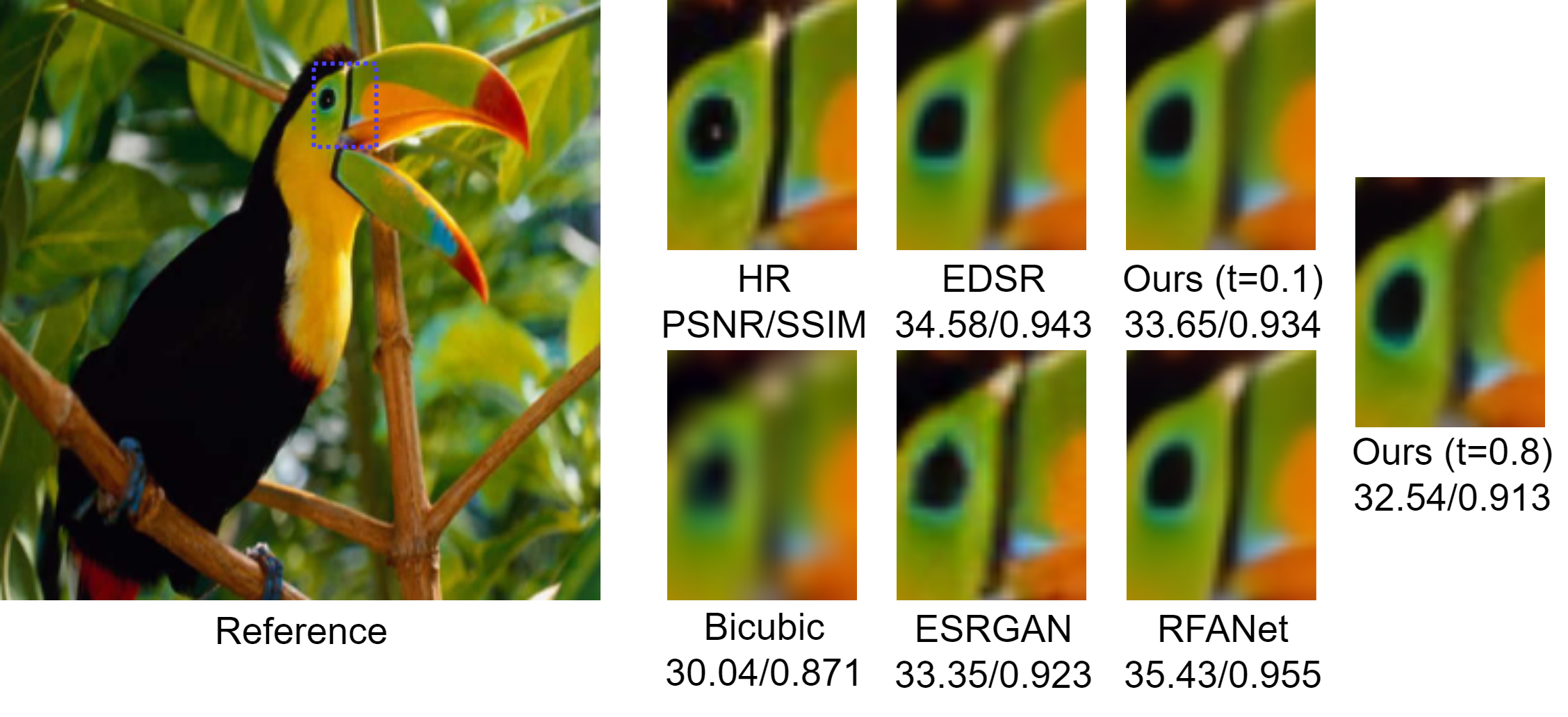}
\caption{SR output comparison between multiple models for the bird picture of the Set5 DataSet.}
\label{fig:bird_comp}
\end{figure}

\cref{fig:bulls_comp,fig:soldier_comp,fig:bird_comp} show a visual comparison of two pictures from BSD100 dataset between the original HR image, Bicubic, EDSR, ESRGAN, RFANet and our method with both $0.1$ and $0.8$ temperatures.

It can be observed that the points made in the quantitative section still stand, as EDSR, having the best PSNR and SSIM scores, has a smoother and blurrier look, and our model with 0.1 temperature looks closer to it. As for the model with 0.8 temperature, it introduces more details compared to EDSR. It is still blurrier than the outputs of ESRGAN but has fewer artifacts and it is able to reproduce some details without introducing any generative noise. As for the RFANet, the images are still blurrier, but having more visual similarities with our method than EDSR. For this reason in most metrics it gets a better score, but visually it still does not generate highly detailed features.

In \cref{fig:bulls_comp,fig:soldier_comp} it can be observed that ESRGAN produces some artifacts on the bull's head and the person's hand, while our model retains the structure of the objects. In \cref{fig:bird_comp} we can again see how the eye of the bird has a different shape and a more averaged-out look in the case of the EDSR, and even more drastic shape change in the case of the ESRGAN, while our models keep the rounder shape, while not averaging out the outer colors as much.

\subsubsection{Temperature.}
\begin{figure}
\centering
\includegraphics[height=5.5cm]{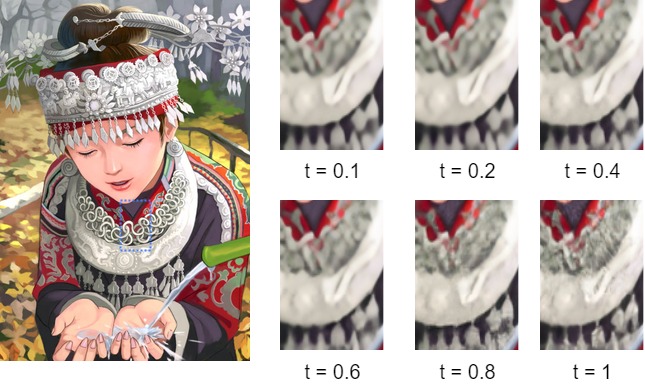}
\caption{Prior sampling difference with varying temperature values for 256x256 images (comic picture from Set14 dataset).}
\label{fig:temp_256}
\end{figure}

\begin{figure}
\centering
\includegraphics[height=4.32cm]{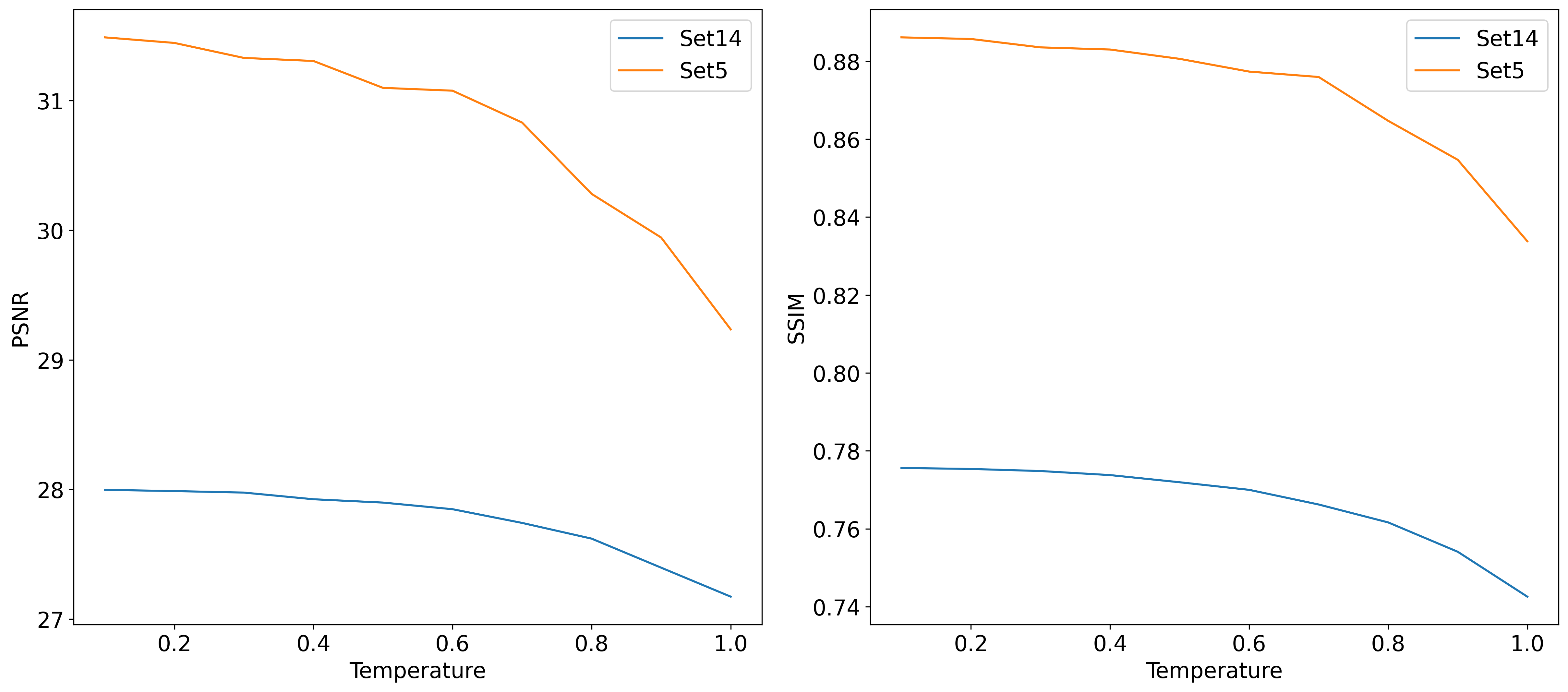}
\caption{PSNR and SSIM scores of Prior samples with varying temperature values for Set5 and Set14 datasets.}
\label{fig:temp_PSNR}
\end{figure}

The ``temperature'' parameter $t$, taking values between 0 and 1, is used in VDVAE when sampling from prior in generative mode, often resulting in higher-quality samples when lowered as observed in previous work \cite{kingma2018glow,vahdat2020nvae}.
Reducing the temperature results in reducing the variance of the Gaussian distributions in the prior and so achieving more regularity in the generated samples. \cref{fig:temp_256} shows examples of samples with different temperatures. We can observe how samples taken with a lower temperature look smoother, whereas those taken with a higher temperature have more details but also more artifacts. 
We corroborate this quantitatively in \cref{fig:temp_PSNR}, which shows that the PSNR and SSIM scores (for Set5 and Set14) both decrease as the sampling temperature is increased. This agrees with our qualitative observations, as PSNR and SSIM measures are usually higher for images that are more averaged out and contain less noise.

\subsubsection{Patch Size.}
\begin{figure}
\centering
\includegraphics[height=4.4cm]{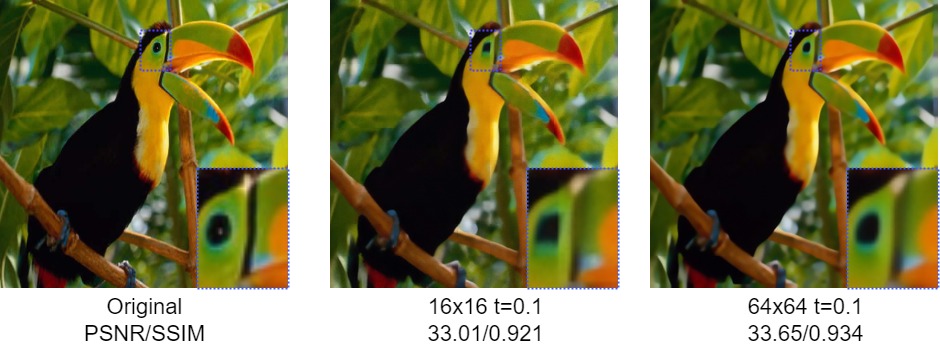}
\caption{16x16 and 64x64 patch size model outputs for a Set5 bird image.}
\label{fig:16vs64patch}
\end{figure}

A crucial parameter in our super-resolution method is the size of patches to which we apply super-resolution addressed also in \cite{zhang2018residual,sajjadi2017enhancenet}. After experimenting with patches of size 16x16 and 64x64 (i.e., 64x64 and 256x256 after super-resolution), we observed that the 16x16 patch size models were generally performing worse than their counterparts with bigger patch sizes, both in terms of PSNR and SSIM, and in a perceptual sense as the models fail to recreate details that the 64x64 patch models have no problem with. This can also be seen in \cref{fig:16vs64patch}, especially on the bird's eye, as the general shape and sharpness cannot be recreated by the 16x16 patch size model.

We hypothesise that as the patch size gets smaller, the amount of details found in a patch becomes lesser, and the models will not be able to recreate those details anymore based on context, as the patches will start to look more similar to each other and generic.

\subsubsection{Activations only in posterior.}
\begin{figure}[!h]
\centering
\includegraphics[height=2.8cm]{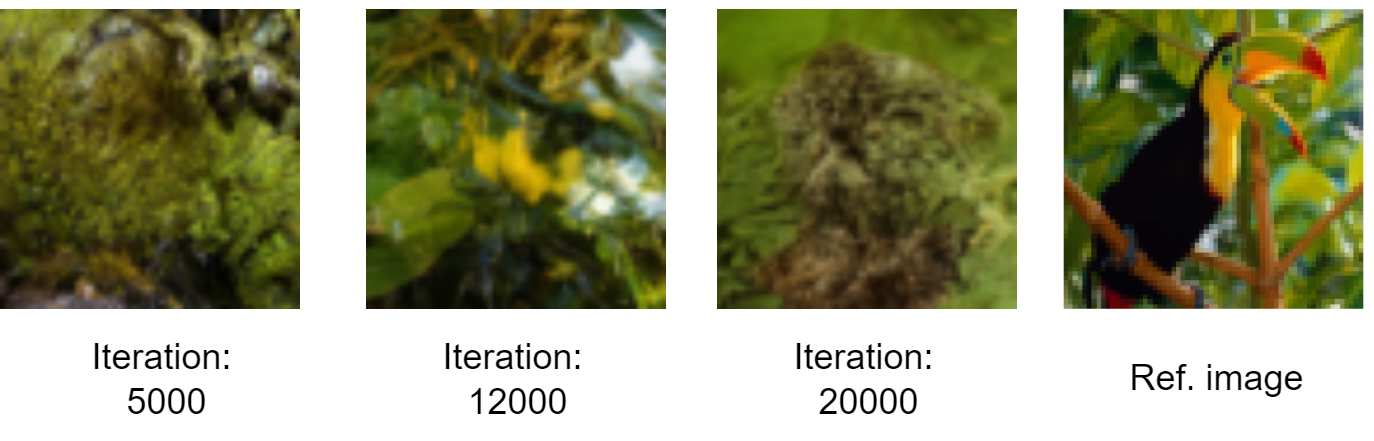}
\caption{The first three images are test samples taken during the training process, while the forth image is the reference.}
\label{fig:acts_post}
\end{figure}

\begin{figure}
\centering
\includegraphics[height=8cm]{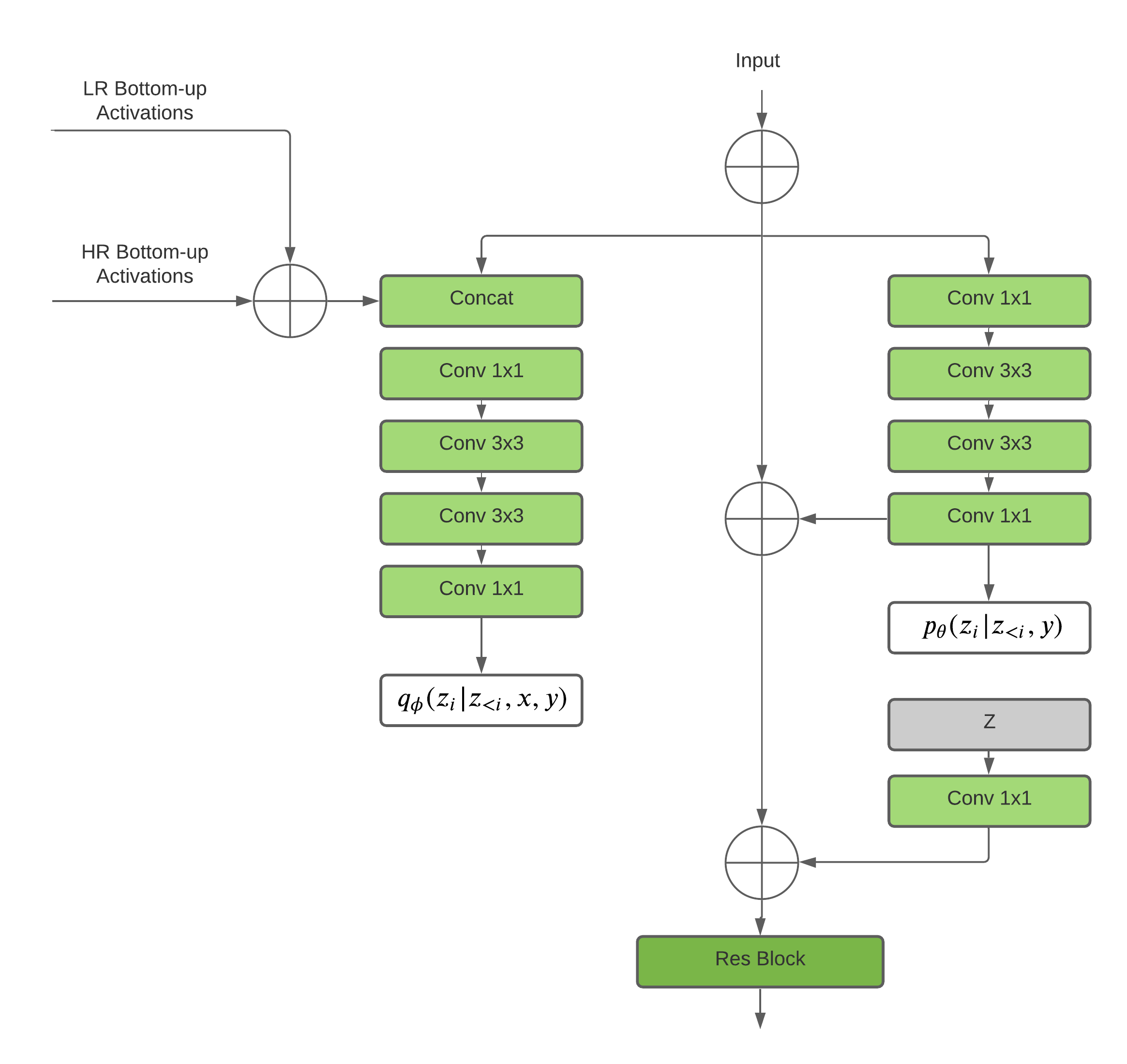}
\caption{Top-down block adding activations in posterior.}
\label{fig:top_down_post}
\end{figure}

As another ablation study, we investigated the scenario where the activations from the \LRenc are passed only in the posterior part of the top-down block as shown on \cref{fig:top_down_post}. Doing only this, the network does not get enough information during the learning phase, only being able to generate more global features of the images, without any fine details as observed in \cref{fig:acts_post}.

\section{Conclusions}

In this paper, we investigated the use of Very Deep Variational Autoencoders (VDVAE) for the purpose of generating super-resolution (SR) images. After the introduction of the proposed VDVAE-SR model, and based on the results presented, we conclude that the introduced model and its quantitative and qualitative results are satisfying as they are comparable to other popular methods, generating images that compensate between image sharpness and visual artifacts. As being part of the scarce family of VAE-based models for image super-resolution and the first to our knowledge that uses a deep hierarchical architecture, we believe that our proposed method still has a lot of space for building upon, to improve the results even further, as multiple modifications such as changes to training time, layer architecture, or the use of more flexible distributions can be investigated in the future.

\clearpage
\bibliographystyle{splncs04}
\bibliography{egbib}

\begin{thebibliography}{10}
\providecommand{\url}[1]{\texttt{#1}}
\providecommand{\urlprefix}{URL }
\providecommand{\doi}[1]{https://doi.org/#1}

\bibitem{Agustsson_2017_CVPR_Workshops}
Agustsson, E., Timofte, R.: Ntire 2017 challenge on single image
  super-resolution: Dataset and study. In: The IEEE Conference on Computer
  Vision and Pattern Recognition (CVPR) Workshops (7 2017)

\bibitem{bachlechner2020rezero}
Bachlechner, T., Majumder, B.P., Mao, H.H., Cottrell, G.W., McAuley, J.: Rezero
  is all you need: Fast convergence at large depth (2020)

\bibitem{BMVC.26.135}
Bevilacqua, M., Roumy, A., Guillemot, C., line Alberi~Morel, M.: Low-complexity
  single-image super-resolution based on nonnegative neighbor embedding. In:
  Proceedings of the British Machine Vision Conference. pp. 135.1--135.10. BMVA
  Press (2012). \doi{http://dx.doi.org/10.5244/C.26.135}

\bibitem{brock2018large}
Brock, A., Donahue, J., Simonyan, K.: Large scale gan training for high
  fidelity natural image synthesis. arXiv preprint arXiv:1809.11096  (2018)

\bibitem{pixel_snail}
Chen, X., Mishra, N., Rohaninejad, M., Abbeel, P.: {P}ixel{SNAIL}: An improved
  autoregressive generative model. In: Dy, J., Krause, A. (eds.) Proceedings of
  the 35th International Conference on Machine Learning. Proceedings of Machine
  Learning Research, vol.~80, pp. 864--872. PMLR (10--15 Jul 2018),
  \url{https://proceedings.mlr.press/v80/chen18h.html}

\bibitem{child2020very}
Child, R.: Very deep vaes generalize autoregressive models and can outperform
  them on images. arXiv preprint arXiv:2011.10650  (2020)

\bibitem{Dai_2019_CVPR}
Dai, T., Cai, J., Zhang, Y., Xia, S.T., Zhang, L.: Second-order attention
  network for single image super-resolution. In: Proceedings of the IEEE/CVF
  Conference on Computer Vision and Pattern Recognition (CVPR) (June 2019)

\bibitem{dhariwal2021diffusion}
Dhariwal, P., Nichol, A.: Diffusion models beat gans on image synthesis. CoRR
  \textbf{abs/2105.05233} (2021), \url{https://arxiv.org/abs/2105.05233}

\bibitem{ding2020iqa}
Ding, K., Ma, K., Wang, S., Simoncelli, E.P.: Image quality assessment:
  Unifying structure and texture similarity. CoRR  \textbf{abs/2004.07728}
  (2020), \url{https://arxiv.org/abs/2004.07728}

\bibitem{dong2015image}
Dong, C., Loy, C.C., He, K., Tang, X.: Image super-resolution using deep
  convolutional networks. IEEE transactions on pattern analysis and machine
  intelligence  \textbf{38}(2),  295--307 (2015)

\bibitem{gatopoulos2020super}
Gatopoulos, I., Stol, M., Tomczak, J.M.: Super-resolution variational
  auto-encoders. arXiv preprint arXiv:2006.05218  (2020)

\bibitem{goodfellow2014generative}
Goodfellow, I., Pouget-Abadie, J., Mirza, M., Xu, B., Warde-Farley, D., Ozair,
  S., Courville, A., Bengio, Y.: Generative adversarial nets. Advances in
  neural information processing systems  \textbf{27} (2014)

\bibitem{he2016deep}
He, K., Zhang, X., Ren, S., Sun, J.: Deep residual learning for image
  recognition. In: Proceedings of the IEEE conference on computer vision and
  pattern recognition. pp. 770--778 (2016)

\bibitem{ddpm}
Ho, J., Jain, A., Abbeel, P.: Denoising diffusion probabilistic models.
  Advances in Neural Information Processing Systems  \textbf{33},  6840--6851
  (2020)

\bibitem{cascaded_ddpm}
Ho, J., Saharia, C., Chan, W., Fleet, D.J., Norouzi, M., Salimans, T.: Cascaded
  diffusion models for high fidelity image generation. Journal of Machine
  Learning Research  \textbf{23}(47),  1--33 (2022)

\bibitem{Huang_2015_CVPR}
Huang, J.B., Singh, A., Ahuja, N.: Single image super-resolution from
  transformed self-exemplars. In: Proceedings of the IEEE Conference on
  Computer Vision and Pattern Recognition (CVPR) (6 2015)

\bibitem{hyun2020varsr}
Hyun, S., Heo, J.P.: Varsr: Variational super-resolution network for very low
  resolution images. In: European Conference on Computer Vision. pp. 431--447.
  Springer (2020)

\bibitem{jolicoeur2018relativistic}
Jolicoeur-Martineau, A.: The relativistic discriminator: a key element missing
  from standard gan. arXiv preprint arXiv:1807.00734  (2018)

\bibitem{karras2019style}
Karras, T., Laine, S., Aila, T.: A style-based generator architecture for
  generative adversarial networks. In: Proceedings of the IEEE/CVF conference
  on computer vision and pattern recognition. pp. 4401--4410 (2019)

\bibitem{kingma2014adam}
Kingma, D.P., Ba, J.: Adam: A method for stochastic optimization. arXiv
  preprint arXiv:1412.6980  (2014)

\bibitem{kingma2013auto}
Kingma, D.P., Welling, M.: Auto-encoding variational bayes. arXiv preprint
  arXiv:1312.6114  (2013)

\bibitem{kingma2018glow}
Kingma, D.P., Dhariwal, P.: Glow: Generative flow with invertible 1x1
  convolutions. Advances in neural information processing systems  \textbf{31}
  (2018)

\bibitem{kingma2016improved}
Kingma, D.P., Salimans, T., Jozefowicz, R., Chen, X., Sutskever, I., Welling,
  M.: Improved variational inference with inverse autoregressive flow. Advances
  in neural information processing systems  \textbf{29} (2016)

\bibitem{ledig2017photo}
Ledig, C., Theis, L., Husz{\'a}r, F., Caballero, J., Cunningham, A., Acosta,
  A., Aitken, A., Tejani, A., Totz, J., Wang, Z., et~al.: Photo-realistic
  single image super-resolution using a generative adversarial network. In:
  Proceedings of the IEEE conference on computer vision and pattern
  recognition. pp. 4681--4690 (2017)

\bibitem{sr_diff}
Li, H., Yang, Y., Chang, M., Chen, S., Feng, H., Xu, Z., Li, Q., Chen, Y.:
  Srdiff: Single image super-resolution with diffusion probabilistic models.
  Neurocomputing  (2022)

\bibitem{lim2017enhanced}
Lim, B., Son, S., Kim, H., Nah, S., Mu~Lee, K.: Enhanced deep residual networks
  for single image super-resolution. In: Proceedings of the IEEE conference on
  computer vision and pattern recognition workshops. pp. 136--144 (2017)

\bibitem{9156371}
Liu, J., Zhang, W., Tang, Y., Tang, J., Wu, G.: Residual feature aggregation
  network for image super-resolution. In: 2020 IEEE/CVF Conference on Computer
  Vision and Pattern Recognition (CVPR). pp. 2356--2365 (2020).
  \doi{10.1109/CVPR42600.2020.00243}

\bibitem{maaloe2019biva}
Maal{\o}e, L., Fraccaro, M., Li{\'e}vin, V., Winther, O.: Biva: A very deep
  hierarchy of latent variables for generative modeling. arXiv preprint
  arXiv:1902.02102  (2019)

\bibitem{937655}
Martin, D., Fowlkes, C., Tal, D., Malik, J.: A database of human segmented
  natural images and its application to evaluating segmentation algorithms and
  measuring ecological statistics. In: Proceedings Eighth IEEE International
  Conference on Computer Vision. ICCV 2001. vol.~2, pp. 416--423 vol.2 (2001)

\bibitem{Matsui_2016}
Matsui, Y., Ito, K., Aramaki, Y., Fujimoto, A., Ogawa, T., Yamasaki, T.,
  Aizawa, K.: Sketch-based manga retrieval using manga109 dataset. Multimedia
  Tools and Applications  \textbf{76}(20),  21811–21838 (11 2016)

\bibitem{nichol2021improved}
Nichol, A.Q., Dhariwal, P.: Improved denoising diffusion probabilistic models.
  In: Meila, M., Zhang, T. (eds.) Proceedings of the 38th International
  Conference on Machine Learning. Proceedings of Machine Learning Research,
  vol.~139, pp. 8162--8171. PMLR (18--24 Jul 2021),
  \url{https://proceedings.mlr.press/v139/nichol21a.html}

\bibitem{niu2020single}
Niu, B., Wen, W., Ren, W., Zhang, X., Yang, L., Wang, S., Zhang, K., Cao, X.,
  Shen, H.: Single image super-resolution via a holistic attention network. In:
  European conference on computer vision. pp. 191--207. Springer (2020)

\bibitem{pixel_cnn}
van~den Oord, A., Kalchbrenner, N., Espeholt, L., kavukcuoglu, k., Vinyals, O.,
  Graves, A.: Conditional image generation with pixelcnn decoders. In: Lee, D.,
  Sugiyama, M., Luxburg, U., Guyon, I., Garnett, R. (eds.) Advances in Neural
  Information Processing Systems. vol.~29. Curran Associates, Inc. (2016),
  \url{https://proceedings.neurips.cc/paper/2016/file/b1301141feffabac455e1f90a7de2054-Paper.pdf}

\bibitem{pixel_rnn}
Oord, A.V., Kalchbrenner, N., Kavukcuoglu, K.: Pixel recurrent neural networks.
  In: Balcan, M.F., Weinberger, K.Q. (eds.) Proceedings of The 33rd
  International Conference on Machine Learning. Proceedings of Machine Learning
  Research, vol.~48, pp. 1747--1756. PMLR, New York, New York, USA (20--22 Jun
  2016), \url{https://proceedings.mlr.press/v48/oord16.html}

\bibitem{image-transformer}
Parmar, N., Vaswani, A., Uszkoreit, J., Kaiser, L., Shazeer, N., Ku, A., Tran,
  D.: Image transformer. In: Dy, J., Krause, A. (eds.) Proceedings of the 35th
  International Conference on Machine Learning. Proceedings of Machine Learning
  Research, vol.~80, pp. 4055--4064. PMLR (10--15 Jul 2018),
  \url{https://proceedings.mlr.press/v80/parmar18a.html}

\bibitem{Pisharoty}
Pisharoty, N., Jadhav, M., Dandawate, Y.: Performance evaluation of structural
  similarity index metric in different colorspaces for hvs based assessment of
  quality of colour images. International Journal of Engineering and Technology
   \textbf{5},  1555--1562 (04 2013)

\bibitem{rezende2014stochastic}
Rezende, D.J., Mohamed, S., Wierstra, D.: Stochastic backpropagation and
  approximate inference in deep generative models. In: International conference
  on machine learning. pp. 1278--1286. PMLR (2014)

\bibitem{sajjadi2017enhancenet}
Sajjadi, M.S., Scholkopf, B., Hirsch, M.: Enhancenet: Single image
  super-resolution through automated texture synthesis. In: Proceedings of the
  IEEE international conference on computer vision. pp. 4491--4500 (2017)

\bibitem{NIPS2015_8d55a249}
Sohn, K., Lee, H., Yan, X.: Learning structured output representation using
  deep conditional generative models. In: Cortes, C., Lawrence, N., Lee, D.,
  Sugiyama, M., Garnett, R. (eds.) Advances in Neural Information Processing
  Systems. vol.~28. Curran Associates, Inc. (2015),
  \url{https://proceedings.neurips.cc/paper/2015/file/8d55a249e6baa5c06772297520da2051-Paper.pdf}

\bibitem{sonderby2016ladder}
S{\o}nderby, C.K., Raiko, T., Maal{\o}e, L., S{\o}nderby, S.K., Winther, O.:
  Ladder variational autoencoders. arXiv preprint arXiv:1602.02282  (2016)

\bibitem{tong2017image}
Tong, T., Li, G., Liu, X., Gao, Q.: Image super-resolution using dense skip
  connections. In: Proceedings of the IEEE international conference on computer
  vision. pp. 4799--4807 (2017)

\bibitem{nade}
Uria, B., C{\^o}t{\'e}, M.A., Gregor, K., Murray, I., Larochelle, H.: Neural
  autoregressive distribution estimation. The Journal of Machine Learning
  Research  \textbf{17}(1),  7184--7220 (2016)

\bibitem{vahdat2020nvae}
Vahdat, A., Kautz, J.: Nvae: A deep hierarchical variational autoencoder. arXiv
  preprint arXiv:2007.03898  (2020)

\bibitem{wang2018esrgan}
Wang, X., Yu, K., Wu, S., Gu, J., Liu, Y., Dong, C., Qiao, Y., Change~Loy, C.:
  Esrgan: Enhanced super-resolution generative adversarial networks. In:
  Proceedings of the European conference on computer vision (ECCV) workshops.
  pp.~0--0 (2018)

\bibitem{zeyde2010single}
Zeyde, R., Elad, M., Protter, M.: On single image scale-up using
  sparse-representations. In: International conference on curves and surfaces.
  pp. 711--730. Springer (2010)

\bibitem{zhang2018residual}
Zhang, Y., Tian, Y., Kong, Y., Zhong, B., Fu, Y.: Residual dense network for
  image super-resolution. In: Proceedings of the IEEE conference on computer
  vision and pattern recognition. pp. 2472--2481 (2018)

\bibitem{zhu2017unpaired}
Zhu, J.Y., Park, T., Isola, P., Efros, A.A.: Unpaired image-to-image
  translation using cycle-consistent adversarial networks. In: Proceedings of
  the IEEE international conference on computer vision. pp. 2223--2232 (2017)

\end{thebibliography}

\appendix

\section{Additional qualitative comparison pictures}
In this section, we show additional qualitative results compared to EDSR \cite{lim2017enhanced} and ESRGAN \cite{wang2018esrgan} on the test set of the BSD100 dataset \cite{937655}. The popular BSD100 dataset consists of images with a broad range of styles ranging from natural images to object-specific ones. We show results of our model with sampling temperature of 0.1, 0.8 and 1. Lower temperature reduces the variance of the Gaussian distributions of the prior, resulting in more averaged-out images, but also reducing noise. We believe that sampling with a temperature of $t = 0.8$ provides a good trade-off between noise reduction and preservation of details.

\begin{figure}
\centering
\includegraphics[height = 17 cm]{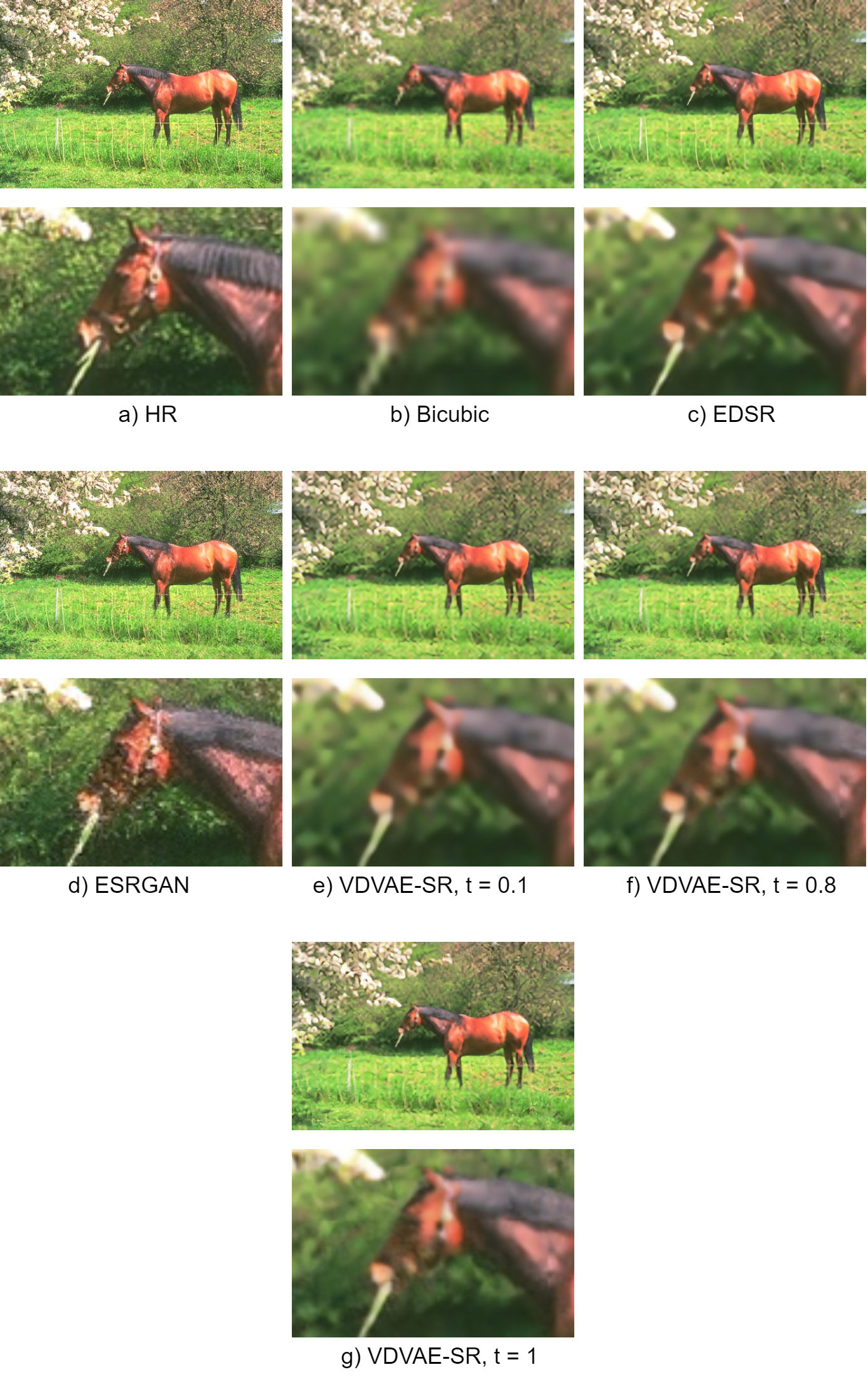}
\caption{SR output comparison between multiple models for picture 291000 of the BSD100 dataset.}
%\label{fig:horse_comp}
\end{figure}

\begin{figure}
\centering
\includegraphics[height = 17 cm]{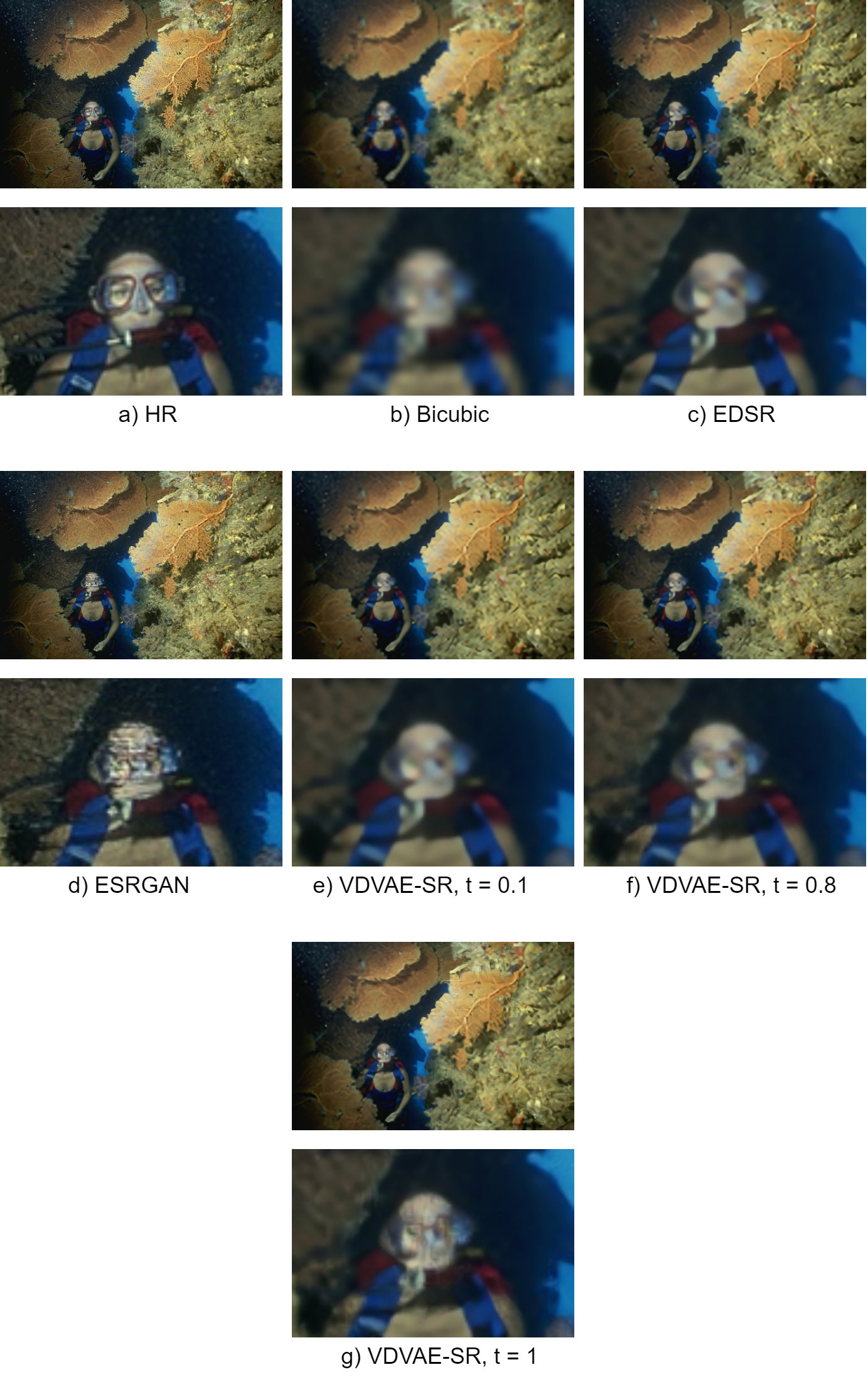}
\caption{SR output comparison between multiple models for picture 156065 of the BSD100 dataset.}
%\label{fig:scuba_comp}
\end{figure}

\begin{figure}
\centering
\includegraphics[height = 17 cm]{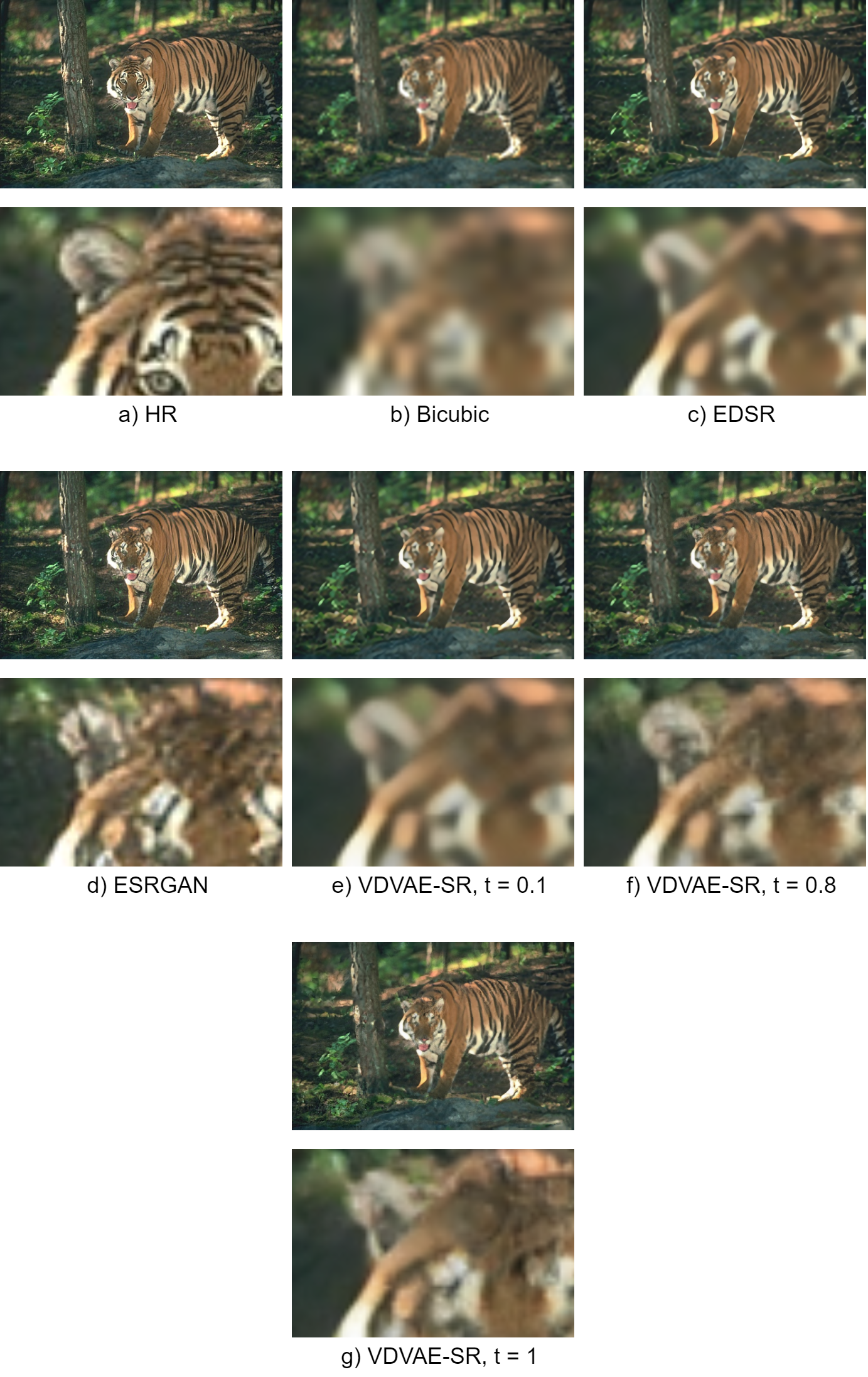}
\caption{SR output comparison between multiple models for picture 108005 of the BSD100 dataset.}
%\label{fig:tiger1_comp}
\end{figure}

\begin{figure}
\centering
\includegraphics[height = 17 cm]{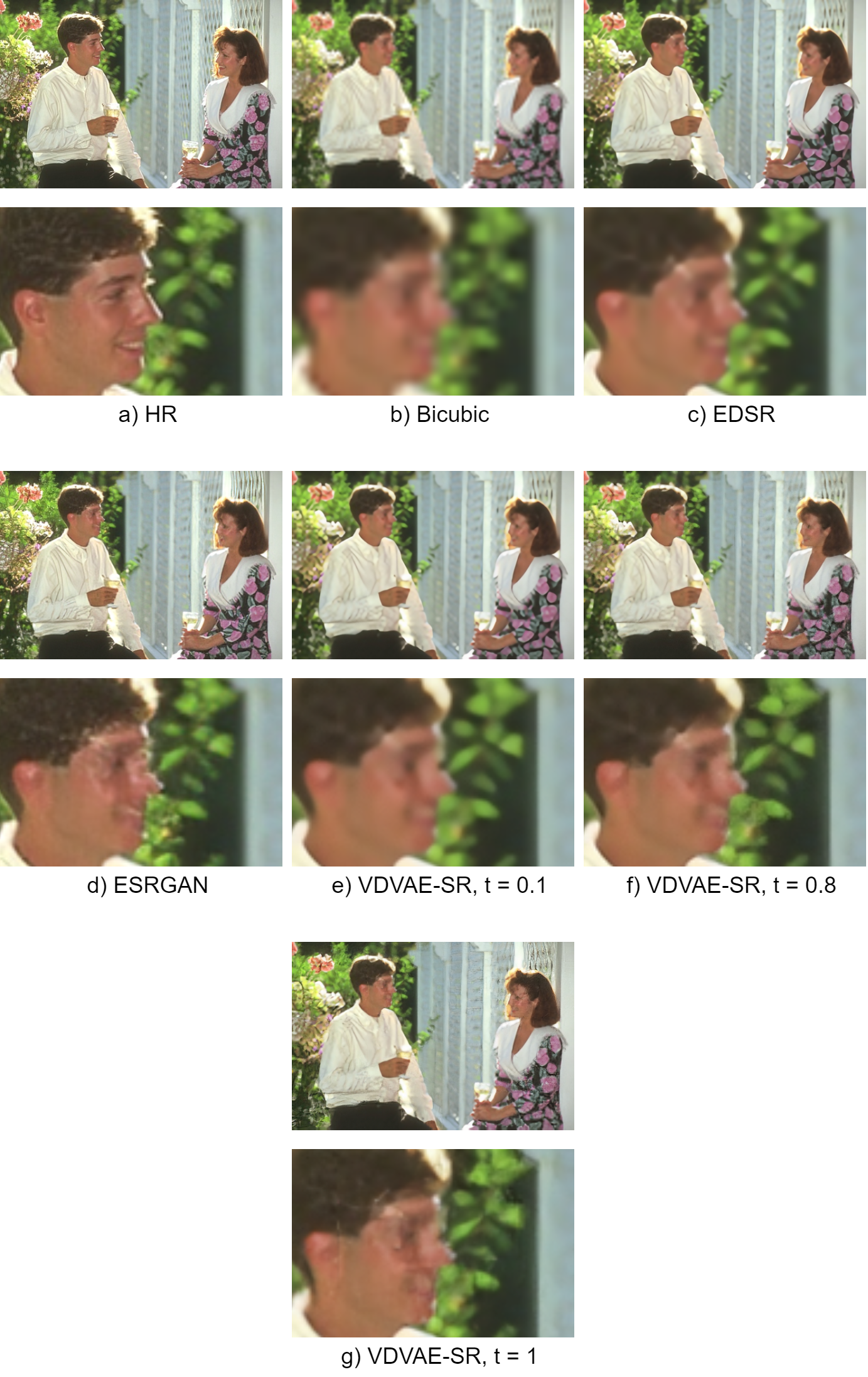}
\caption{SR output comparison between multiple models for picture 157055 of the BSD100 dataset.}
%\label{fig:manandwoman_comp}
\end{figure}

\begin{figure}
\centering
\includegraphics[height = 17 cm]{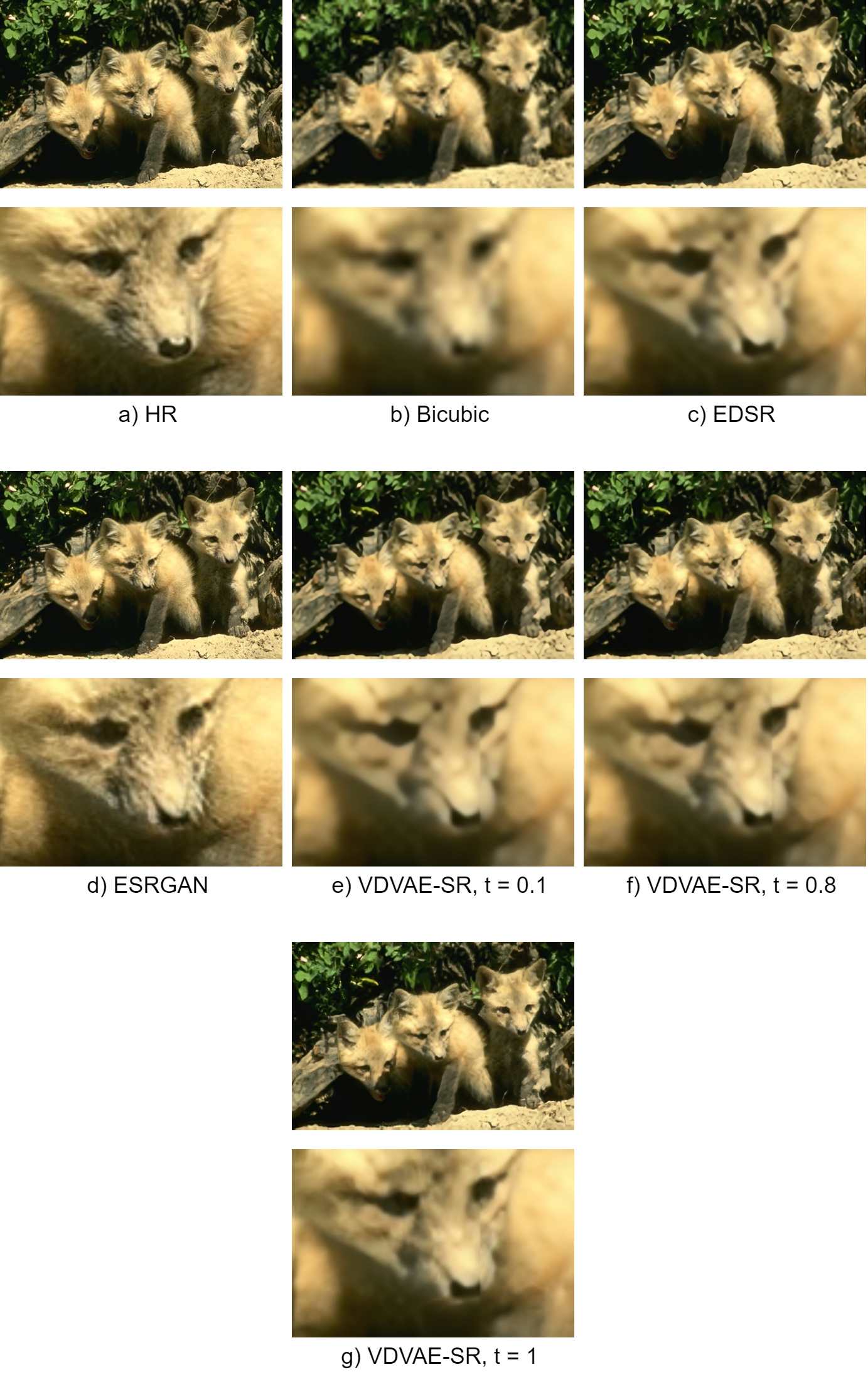}
\caption{SR output comparison between multiple models for picture 159008 of the BSD100 dataset.}
%\label{fig:foxes_comp}
\end{figure}

\begin{figure}
\centering
\includegraphics[height = 17 cm]{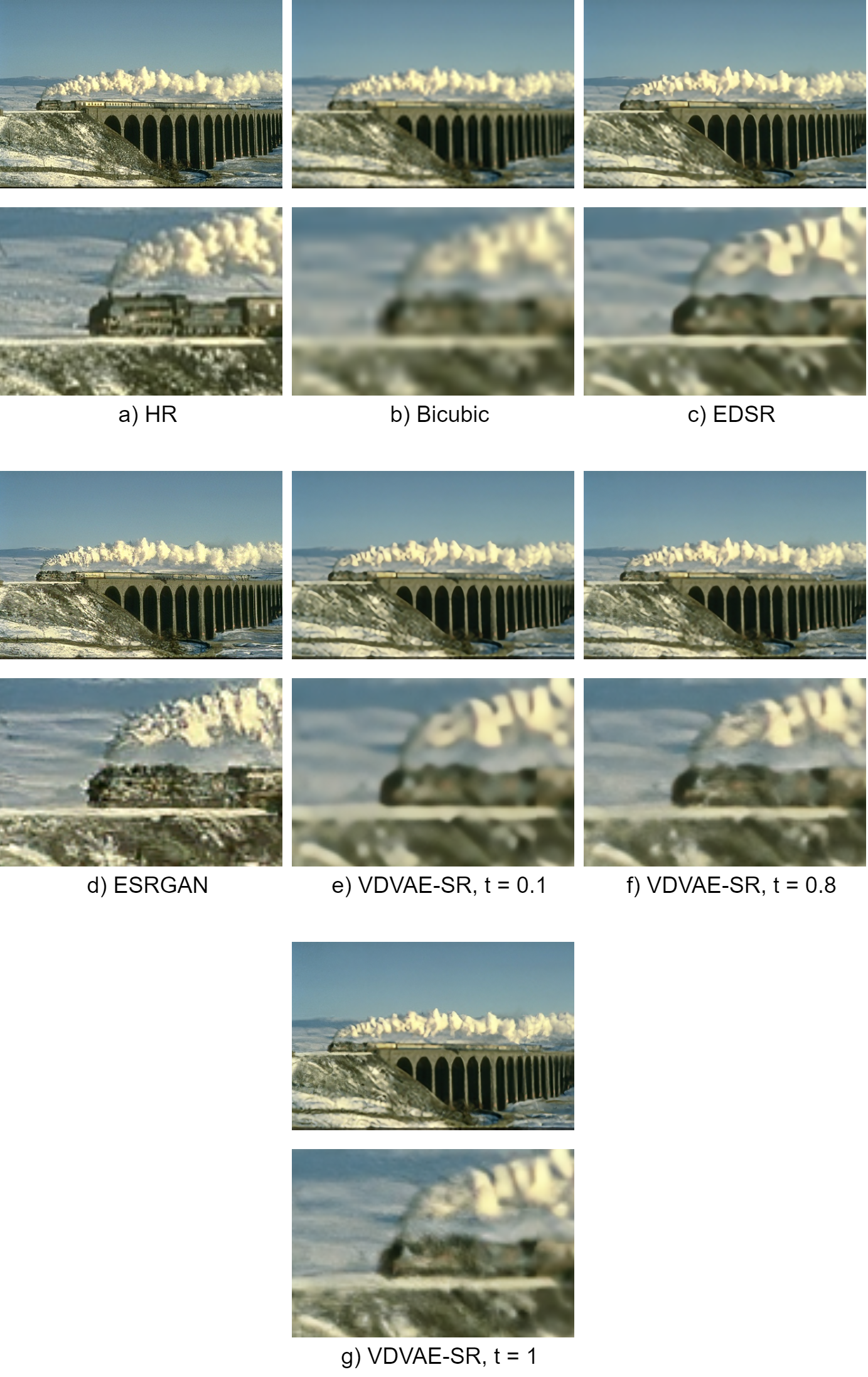}
\caption{SR output comparison between multiple models for picture 182053 of the BSD100 dataset.}
%\label{fig:train_comp}
\end{figure}

\begin{figure}
\centering
\includegraphics[height = 17 cm]{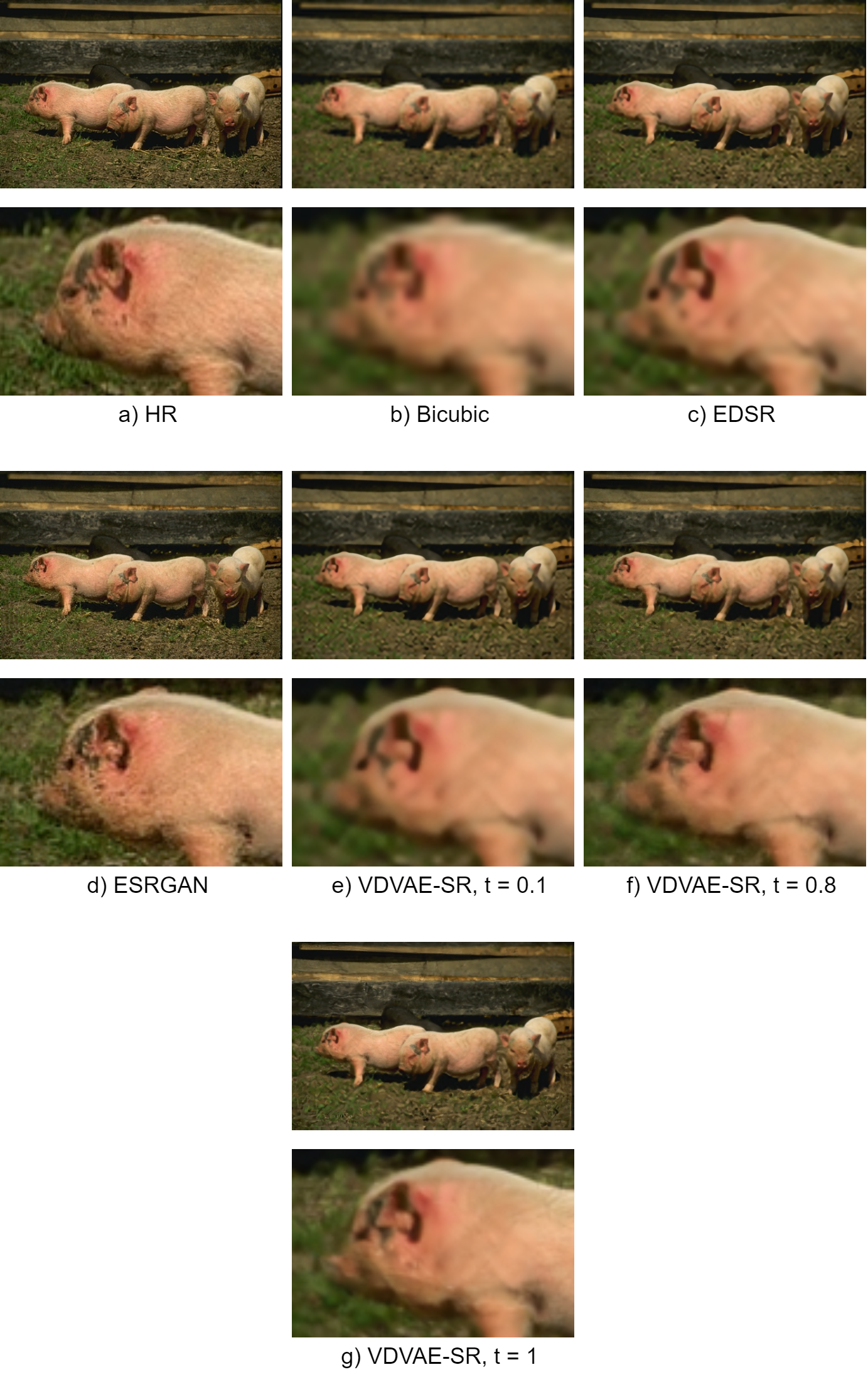}
\caption{SR output comparison between multiple models for picture 66053 of the BSD100 dataset.}
%\label{fig:pigs_comp}
\end{figure}

\begin{figure}
\centering
\includegraphics[height = 17 cm]{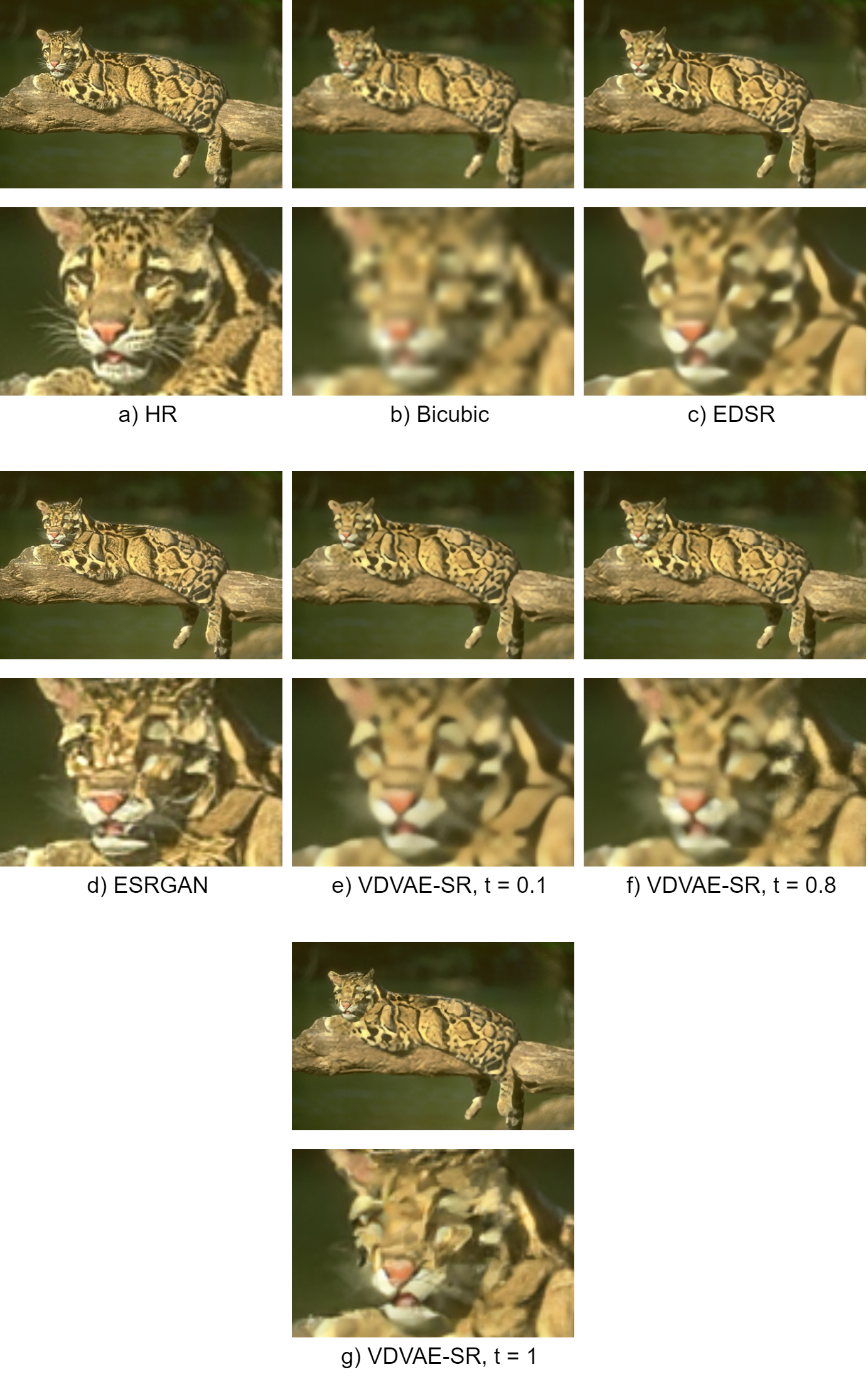}
\caption{SR output comparison between multiple models for picture 160068 of the BSD100 dataset.}
%\label{fig:pigs_comp}
\end{figure}

\begin{figure}
\centering
\includegraphics[height = 17 cm]{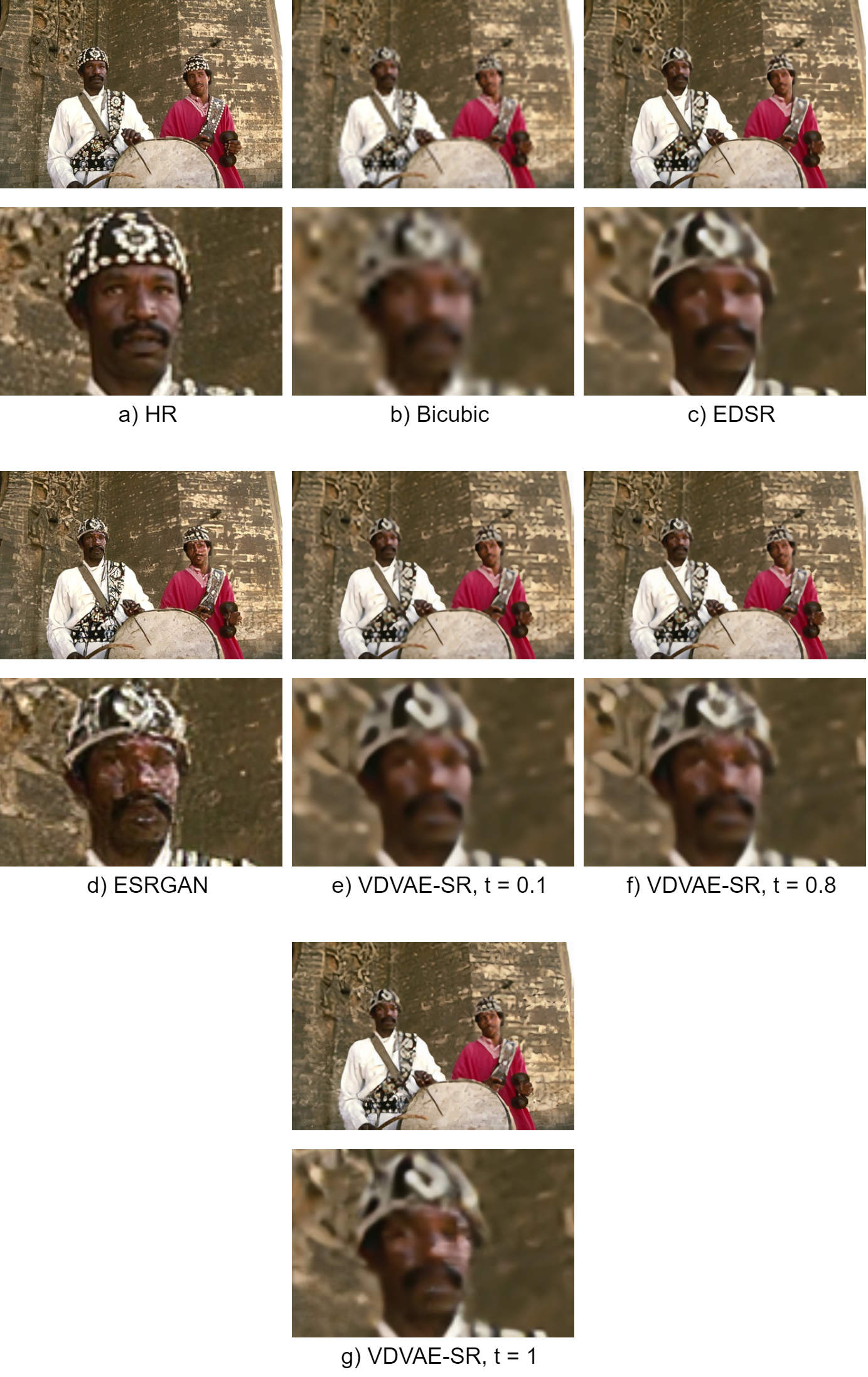}
\caption{SR output comparison between multiple models for picture 229036 of the BSD100 dataset.}
%\label{fig:pigs_comp}
\end{figure}

\begin{figure}
\centering
\includegraphics[height = 17 cm]{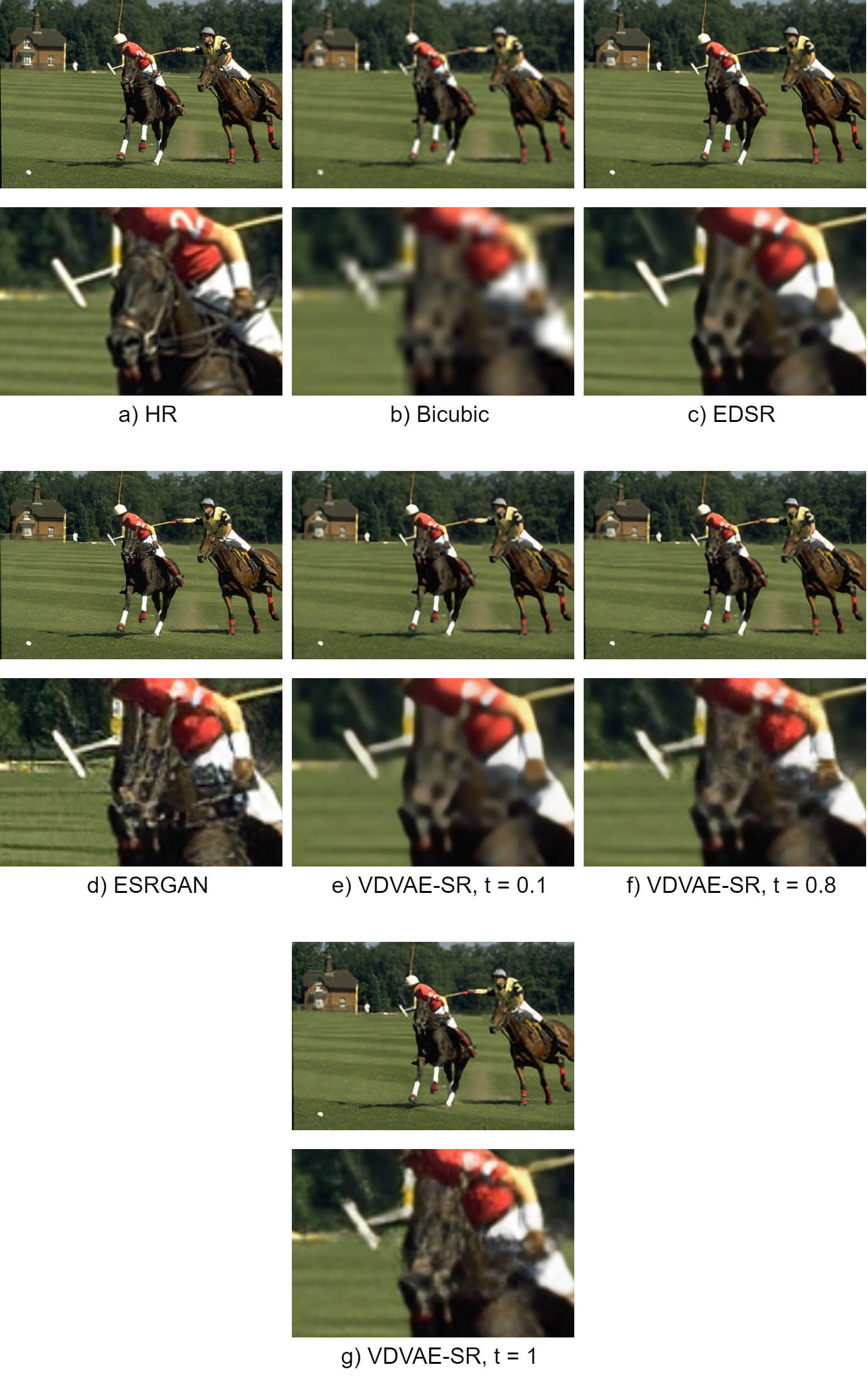}
\caption{SR output comparison between multiple models for picture 361010 of the BSD100 dataset.}
%\label{fig:pigs_comp}
\end{figure}

\end{document}